\documentclass[11pt]{article}

\usepackage[final]{acl}
\usepackage{times}
\usepackage{latexsym}

\usepackage[T1]{fontenc}
\usepackage[utf8]{inputenc}

\usepackage{microtype}
\usepackage{inconsolata}
\usepackage{graphicx}
\usepackage{array}
\usepackage{amsmath}
\usepackage{authblk}
\usepackage{amssymb}
\usepackage{parskip}

\usepackage{algorithm}
\usepackage{algpseudocode}
\usepackage{caption}
\usepackage{amsmath}
\usepackage{booktabs}
\usepackage{multirow}
\usepackage{adjustbox}
\usepackage{subcaption}
\usepackage{listings}
\usepackage{makecell}
\usepackage[most]{tcolorbox} 
\usepackage{etoc}
\usepackage{color-edits}
\addauthor{kw}{orange}
\addauthor{sx}{red}
\definecolor{editblue}{RGB}{44,86,138}
\addauthor{hh}{editblue}

\usepackage[nameinlink]{cleveref}

\title{Compute Allocation for Self-Evolving LLMs: From Depth–Breadth to Multi-Armed Bandits}

\newcommand*\samethanks[1][\value{footnote}]{\footnotemark[#1]}

\author{
Sixue Xing\thanks{~~These authors contributed equally to this work.}\textsuperscript{\textrm{1}}\quad
Haoyu He\samethanks[\value{footnote}]\textsuperscript{\textrm{2}}\quad
Kerui Wu\samethanks[\value{footnote}]\textsuperscript{\textrm{3}} \\
Zhuo Yang\textsuperscript{\textrm{4}}\quad
Haozheng Luo\textsuperscript{\textrm{5}}\quad
Tianfan Fu\textsuperscript{\textrm{6, 7}}\quad
Aarthy Nagarajan\textsuperscript{\textrm{1}} \\\\
\textsuperscript{1}University of Notre Dame \quad
\textsuperscript{2}Northeastern University \quad
\textsuperscript{3}University of Massachusetts Amherst \\
\textsuperscript{4}Southeast University \quad
\textsuperscript{5}Northwestern University \quad
\textsuperscript{6}Shanghai Artificial Intelligence Laboratory
\\
\textsuperscript{7}State Key Laboratory for Novel Software Technology at Nanjing University\\
}

\begin{document}
\maketitle

\addtocontents{toc}{\protect\setcounter{tocdepth}{-1}}   


\begin{abstract}
LLM-guided evolutionary search (Evolve systems) has reached state-of-the-art results on mathematical and combinatorial tasks, yet most existing systems report only the best of many runs and leave the run-to-run distribution undocumented. We ask how a fixed budget of LLM calls should be allocated, and how reliably a single run reaches the reported numbers. Sweeping the depth–breadth grid over five models and three tasks, we identify two empirical regularities: a fitness–compute envelope along which capability ordering largely collapses when measured in effective FLOPs, and a bilinear depth–breadth fit with task-specific interaction; both are gated by model–task capability. Motivated by these regularities, we propose BaSE (Bandit-based Self-Evolving), a multi-armed bandit that allocates LLM calls across parallel trajectories. Without changing the model, prompt, or evaluator, BaSE improves mean fitness by 12.3\% over the strongest island-protocol baseline across 8 (model, task) cells, with the largest gains on high-variance settings: a reliability gain from allocation alone. The code is available at: \url{https://github.com/keruiwu/self-evolving-allocation}.
\end{abstract}

\section{Introduction}

\begin{figure}
    \centering
    \includegraphics[width=.95\linewidth]{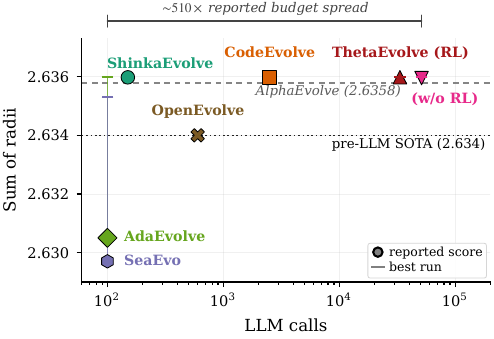}
    \caption{Cost–performance frontier on $n{=}26$ Circle Packing (CP).}
    \label{fig:frontier}
\end{figure}

Large language models are increasingly used as mutation engines for evolutionary search: given a candidate program, a frozen LLM proposes variants; a deterministic evaluator scores them; the best variant seeds the next round \citep{romeraparedes2024funsearch,novikov2025alphaevolve}.
The resulting \emph{Evolve} systems \citep{lange2025shinkaevolve,wang2025thetaevolve,assumpcao2025codeevolve,cemri2026adaevolve} have produced state-of-the-art results across mathematical discovery, combinatorial optimization, and algorithm design. However, the headline numbers \emph{*Evolve} systems report are systematically incomparable: FunSearch reports a 4-of-140 hit rate \citep{romeraparedes2024funsearch}; CodeEvolve displays ``only the best'' \citep{assumpcao2025codeevolve}; AlphaEvolve reports a single number on the $n{=}26$ Circle Packing benchmark \citep{novikov2025alphaevolve}. Reported per-run cost spans more than two orders of magnitude, from $\sim$150 LLM calls in ShinkaEvolve \citep{lange2025shinkaevolve} to the 204{,}800 candidates ThetaEvolve processes in a single run \citep{wang2025thetaevolve}. \Cref{fig:frontier} plots these reports on a common axis: each point is a single hand-picked configuration, and the dominant convention is to report only the best of an unspecified number of runs. 
Few systems report the run-to-run distribution behind these headline numbers. Combined with the order-of-magnitude variation in reported computational cost, this makes the numbers an unreliable guide to single-run performance under realistic deployment settings: existing reports characterize what is achievable on a favorable run, not what a practitioner should expect at a finite computational cost.


The expected performance is shaped by multiple design choices. A stronger base model can generate better mutations; a more informative prompt can guide the search toward more useful edits; and, importantly, allocation determines how the evolutionary process balances exploration and exploitation. This allocation effect is orthogonal to model and prompt quality, and remains meaningful once the model, prompt, and evaluator are fixed. Better allocation can improve expected outcomes by avoiding both premature commitment to weak trajectories and excessive breadth without refinement.

 While \emph{*Evolve} systems are often described as iterative loops that continue until progress saturates, practical deployments must operate under explicit resource constraints, such as a fixed number of model calls, a limited compute allocation, or a bounded experimental campaign. This motivates a controlled study of expected performance and reliability in LLM-guided evolutionary search.
We adopt the fixed-budget perspective \citep{jansen2012fixedbudget} from classical evolutionary computation and bring it to LLM-guided evolution. To this end, we conduct the first empirical study of cost allocation in LLM-guided evolutionary search, characterizing how a fixed number of LLM calls should be spent, and use the resulting picture to ask whether allocation can be precomputed offline or must be discovered online. In summary, our contributions are listed below:

\begin{enumerate}
\item Proposed the first systematic empirical measurement of fixed-budget allocation between exploration and exploitation in LLM-guided evolutionary search.
\item Identified two regularities: \emph{(i)} a performance–compute envelope of attainable fitness against effective FLOPs, along which capability ordering largely collapses among capable models; and \emph{(ii)} a parametric depth–breadth regularity characterizing each model-task cell.
\item Designed an adaptive bandit allocator (\texttt{BaSE}) over
parallel evolutionary trajectories, improving best mean fitness by
\textbf{12.3\%} on average over the strongest island-protocol baseline
across 8 (model, task) cells --- this is not a model improvement, nor a prompt improvement, but an allocation improvement.

\end{enumerate}


\section{Related Work}
\label{sec:related}



\noindent\textbf{LLM-Guided Evolutionary Search.}~
FunSearch \citep{romeraparedes2024funsearch} introduced LLMs as
mutation operators in evolutionary program synthesis, and AlphaEvolve
\citep{novikov2025alphaevolve} scaled the paradigm to
state-of-the-art results. \emph{*Evolve} variants vary individual
axes such as sample efficiency \citep{lange2025shinkaevolve}, test-time
RL \citep{wang2025thetaevolve}, and open-source islands, with adjacent work on code, architectures, and prompts \citep{lehman2022elm,chen2023evoprompting,yang2023opro, xing2026clinicalretrial}. All leave population size and generation count hand-set.
ThetaEvolve only notes qualitatively that small databases progress
faster early but large ones win at scale \citep{wang2025thetaevolve},
and Population-Evolve sweeps population size without holding total
budget fixed \citep{zhang2025populationevolve}. Adjacent allocation
work routes compute between islands \citep{cemri2026adaevolve},
compares evolution to best-of-N and sequential revision
\citep{lee2025evolving}, or focuses on parent sampling
\citep{novikov2025alphaevolve,lange2025shinkaevolve} and
combination operators
\citep{lange2023discovering,meyerson2024language}. ShinkaEvolve \citep{lange2025shinkaevolve} uses a bandit
to ensemble \emph{models} at the final phase; we instead target the
within-run depth--breadth split.

\noindent\textbf{Compute Allocation.}~
Inference-scaling work allocates test-time compute
\citep{snell2024scaling,brown2024monkeys,wu2024inferencescaling,chen2024provablescaling, meir2026more},
with breadth-vs-depth studies on single-query reasoning and tree
search
\citep{sharma2025sequentialedge,wen2025parathinker,inoue2025abmcts,miyamoto2026bgmcts}
reporting empirical, regime-dependent curves; none targets
population-based evolution with parent-conditioned mutation. Our
online allocator draws on multi-armed bandits
\citep{auer2002finite,lattimore2020bandit} and fixed-budget best-arm
identification \citep{audibert2010bestarm,karnin2013almost}, with
precedents in hyperparameter search \citep{li2017hyperband}, prompt
selection \citep{shi2024triple}, MCTS expansion
\citep{inoue2025abmcts}, island routing \citep{cemri2026adaevolve},
and code repair \citep{tang2024coderepair}, all of which
\emph{assume} adaptive allocation helps. Classical EA contributes
optimal-population-under-fixed-compute analyses
\citep{nakano1994population,briesch2023tradeoff}, offspring-size
theory
\citep{jansen2005offspring,doerr2015optimizing,giessen2017interplay,badkobeh2014unbiased},
and the fixed-budget framing
\citep{jansen2012fixedbudget,jansen2014performance}; all assume
random mutation, whereas LLM mutation carries code priors and
discrete attractors that break i.i.d.\ mean-field assumptions, so we
characterize the surface empirically.

\section{Problem Formulation}
\label{sec:approach}

\noindent\textbf{Self-Evolving.}~
At high level, we formulate self-evolution as an iterative optimization process under a fixed inference budget of $C$ LLM calls. Specifically, as described in \Cref{alg:self_evolving}, a Self-Evolve system repeatedly generates a candidate response, evaluates its quality, and refines the proposed solution based on historical response-evaluation pairs.

\begin{algorithm}[ht]
\caption{Self-Evolving Process}
\label{alg:self_evolving}
\footnotesize
\begin{algorithmic}[1]
\Require Task $q$, base LLM $f$, prompt generator $h$, budget (number of LLM calls) $C$

\For{$c = 0, \dots, C-1$}
    \State \label{line:parent_sampling} $\mathrm{prompt} \gets h \left( q, \left(\mathrm{response}^{(j)}, \mathrm{score}^{(j)} \right)_{j \in [c]} \right)$ 
    \State $\mathrm{response}^{(c+1)} \gets f(\mathrm{prompt})$ 
    \State $\mathrm{score}^{(c+1)} \gets \mathrm{Eval} \left( q, \mathrm{response}^{(c+1)} \right)$ 
\EndFor
\State $j^* \gets \arg \max_{j \in [C]}  \mathrm{score}^{(j)}$
\State \Return $\mathrm{response}^{(j^*)}$

\end{algorithmic}
\end{algorithm}

The key design question in this process is the \emph{parent sampling protocol} described in \Cref{line:parent_sampling} of \Cref{alg:self_evolving}: how should the system select and reuse historical responses when constructing the next prompt? This choice determines how the evolutionary process balances exploitation of high-performing solutions with exploration of diverse alternatives. In this work, we explored two representative protocols: \emph{greedy} and \emph{island}:

\noindent\textbf{Greedy Protocol.}~
At each generation, greedy protocol selects the current best-scoring response as the parent and generates $N$ children from it in parallel. Equivalently, it spends $N$ LLM calls per generation refining the best solution found so far.

\noindent\textbf{Island Protocol.}~
The island protocol originates in classical evolutionary computation~\cite{tanese1989distributed} and was brought into LLM-guided evolutionary search by FunSearch~\cite{romeraparedes2024funsearch}, later adopted by many Evolve-style systems; it maintains a population database partitioned into multiple islands. At each generation, it first selects an island according to a MAP-Elites-style coverage rule~\cite{mouret2015illuminating}, and then samples a parent uniformly from that island.


\noindent\textbf{The Depth--Breadth Allocation Problem.}~
Under the greedy protocol, a run of budget $C$ is fully specified by how
that budget is split: $T$ generations of $N$ children each, so that:

\[
  C \;=\; T \cdot N,
\]

The two ends of this split are familiar limits: $T{=}1$ spends the whole
budget on a single generation of parallel samples (best-of-$N$), whereas
$T{=}C$ refines one trajectory for $C$ sequential steps. Let $V(C,T)$ be
the expected best fitness of a run at budget $C$ and depth $T$. Holding the evaluator, prompt template, base
model, and initial program fixed, the split is the only remaining degree
of freedom, and we seek the \emph{compute-optimal depth}


\begin{equation}\label{eq:objective}
\begin{gathered}
  T^{*}(C) = \arg\max_{T} V(C,T), \\
  V_{\max}(C) = \max_{T} V(C,T).
\end{gathered}
\end{equation}

the question we take up in \Cref{sec:scaling_law} is how V(C,T) varies with allocation.


\section{Experiment Setup}
\label{sec:setup-exp}



\noindent\textbf{Tasks.}~ We evaluate three geometric
optimization tasks drawn from AlphaEvolve and shipped with the
OpenEvolve example suite: \textbf{Circle Packing} (CP, $n{=}26$),
\textbf{MinMaxDist} (MMD, $n{=}16$), and \textbf{Heilbronn
Triangle} (HT, $n{=}11$). For each task we use the OpenEvolve
evaluator and initial-program files verbatim, and normalize raw
objectives by the best published construction so that
$\textsc{fitness}{=}1.0$ matches the state-of-the-art. Full problem statements and normalizers are in \Cref{app:task_details}.




\noindent\textbf{Models.}~
We sweep the open-weight Qwen3 family at four sizes: 1.7B, 4B, 8B, and 14B, with thinking mode enabled, and Llama-3.1-8B (Llama). We set the temperature to $0.6$, and top-$p$ to $0.95$. Inference is served by vLLM v0.18 with $\texttt{--quantization fp8}$, $\texttt{--max-model-len}{=}40960$, $\texttt{--max-num-seqs}{=}16$. The vllm server runs with up to 16 in-flight parallel LLM calls on a single H100 GPU.

\noindent\textbf{Bandits Algorithms.}~
In \Cref{sec:MAB_for_trajectory_allocation}, we explore the impact of different runs on fitness score performance and employ Multi-Armed Bandits (MAB) for adaptive trajectory allocation. We implement three classic MAB algorithms, namely, Upper Confidence Bound (UCB) \cite{auer2002nonstochastic}, Exponential-weight algorithm for Exploration and Exploitation with high Probability (EXP3.P) \cite{auer2002nonstochastic}, and Thompson Sampling (Thompson) \cite{thompson1933likelihood, agrawal2012analysis}. We also add two naive baselines\footnote{We emphasize that they are non-adaptive baselines against MAB policies that choose which evolving trajectory to pull instead of choosing a parent inside the evolving process.}, namely, random, which randomly chooses runs at each round, and round robin, which sequentially pulls each run in turn, to isolate the splitting and redistributing effects, which are two components from our proposed cross-trajectory allocator.

\noindent\textbf{Parameters.}~
We sweep greedy on Qwen3 8B/14B and Llama at $C \in \{8, \dots, 512\}$ with $T \in \{1, 2, 4, \dots, C\}$ (smaller Qwen3 1.7B/4B capped at $C{=}128$), 10 seeds per cell; $T{=}1$ recovers best-of-$N$, $T{=}C$ is pure sequential. Island protocols (OpenEvolve, CodeEvolve, ShinkaEvolve) run at $C{=}512$ on the same three models. For each (model, task) cell, Bandit and all protocols run end-to-end 10 times with independent LLM seeds. We report stratified bootstrap standard errors and 95\% CIs over the 1000 resamples, following the \texttt{rliable} evaluation protocol of \citet{agarwal2021deep}.


\textbf{Post-hoc FLOPs accounting.} Counting LLM calls ($C$) is not 
a fair cost axis across models or protocols: different model sizes 
and different prefix-cache hit rates (greedy reuses one parent prefix 
across $N$ siblings; island rewrites the prompt per call) consume 
substantially different FLOPs at the same $C$. Following 
\citet{hoffmann2022chinchilla}, we charge each LLM call:
\[
\textsc{FLOPs}_{\text{call}} = 2\,P_{\text{active}}\,(p_{\text{prompt}} - p_{\text{cached}} + p_{\text{out}}),
\]
where $P_{\text{active}}$ is the active parameter count and 
$p_{\text{prompt}}, p_{\text{cached}}, p_{\text{out}}$ are the prompt, 
prefix-cached, and completion token counts. Per-run 
FLOPs sum over all calls (full definitions are deferred to \Cref{app:flops}).
FLOPs/$C$ varies by under 15\% across $T$ at fixed $C$ within each model ($R^2 \geq 0.94$, pooled linear fit; \Cref{fig:surface_all8}, top row), so we use $C$ within models and effective FLOPs across models.

\section{Depth--Breadth Allocation}
\label{sec:scaling_law}

We use greedy runs with a fixed budget ($C = T \cdot N$), leaving the allocation between depth and breadth as the only remaining degree of freedom.
We analyze the full $(C,T)$ sweep in two steps:
\Cref{sec:envelope} asks \emph{how much} fitness a budget buys, on a cost
axis comparable across model sizes. \Cref{sec:production_law} asks
\emph{how} the budget should be split, and fits a parametric regularity to the
depth--breadth surface.

\subsection{The Fitness-Compute Envelope}
\label{sec:envelope}

\begin{figure}[t]
    \centering
    \includegraphics[width=\linewidth]{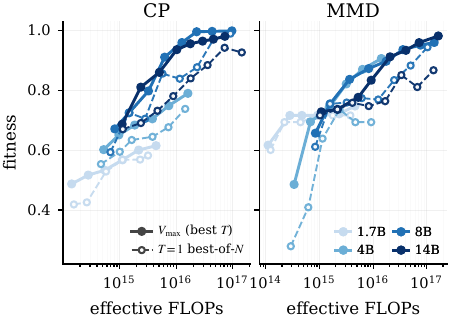}
    \caption{Fitness versus effective FLOPs.  Solid lines report
    $V_{\max}$ over swept depth-breadth allocations; dashed lines report the pure best-of-$N$ baseline ($T=1$). Vertical gap between solid and dashed curves is the gain of 'evolving' from using evaluator feedback across generations rather than spending the full budget on one generation of
    parallel samples.}
    \label{fig:flops_envelope}
\end{figure}


At each budget we summarize the sweep by the best fitness 
$V_{\max}(C)$, the highest fitness achieved 
across all tested depth--breadth allocations at that compute level, 
following the compute-performance envelope convention of 
\citet{hoffmann2022chinchilla}, and study how it scales 
with effective FLOPs.
We further measure compute in effective FLOPs for comparing this envelope across models (\Cref{sec:setup-exp}) and study $V_{\max}$ (best in-sweep allocation) on that axis. Figure~\ref{fig:flops_envelope} reports $V_{\max}$ and the best-of-$N$ baseline against this axis. For every model $V_{\max}$ rises
smoothly and monotonically with compute envelope. This is a clean envelope with no phase transitions or reversals.


\paragraph{Capability ordering collapsing.}
At equal LLM call count, larger models lead, but the lead is not free: a larger model spends proportionally more FLOPs per call. Re-priced in effective FLOPs (\Cref{fig:flops_envelope}), the ordering largely dissolves: on MMD the 4B, 8B and 14B envelopes nearly coincide ($R^2=0.94$); on CP: 8B and 14B coincide over the sub-ceiling range ($R^2=0.93$). This
collapse itself is gated by model-task capability (\Cref{app:effect_model_size}).

\paragraph{Allocation improves fitness.}
The vertical gap between the solid envelope and the dashed best-of-$N$
baseline in \Cref{fig:flops_envelope} is the gain from multi-generation
refinement. At $C{=}128$ it reaches up to a tenth of the normalized fitness
range ($+0.119$ on CP~8B, $+0.102$ on MMD~8B); it persists on unsaturated
tasks ($+0.115$ on MMD~14B at $C{=}512$) but diminishes where the task
saturates ($+0.016$ on MMD~8B). A selection-free test confirms a real onset on MMD but no sharp threshold on CP.
Allocation thus yields real but task-dependent gains, 
whose structure we characterize in \Cref{sec:production_law}.

\begin{figure*}[ht]
    \centering
    \includegraphics[width=.9\linewidth]{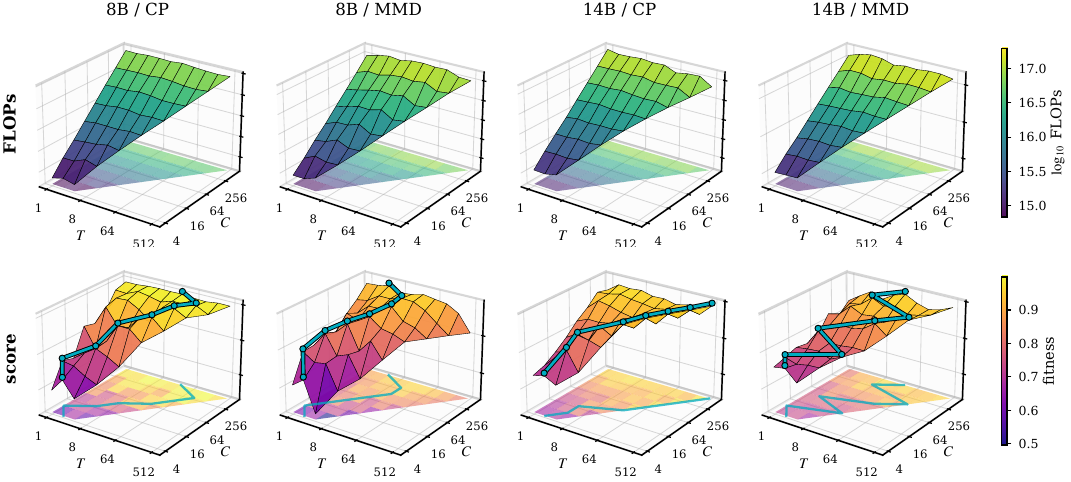}
    \caption{Full $(C,T)$ surfaces for the C=512 sweeps.  Top: effective
    FLOPs; bottom row: mean fitness.  The FLOPs surfaces are nearly flat in
    $T$ at fixed $C$, so changes along the depth axis mostly reflect
    allocation effects rather than hidden cost differences.  The score
    surfaces reveal different allocation geometry: Circle Packing is
    depth-favored with broad plateaus, whereas MinMaxDist has an interior
    ridge.}
    \label{fig:surface_all8}
\end{figure*}

\subsection{The Depth--Breadth Regularity}
\label{sec:production_law}

\begin{table}[t]
    \centering
    \small
    \caption{Posthoc depth--breadth gap model on sub-ceiling cells of the
    C=512 sweeps.  Coefficients are for \Cref{eq:bi-law} with natural logs.
    MMD has a clearly negative interaction (interior ridge) on
    both capable models; CP and HT sit near the corner limit
    ($|c|<0.03$).}
    \label{tab:production_law}
    \begin{adjustbox}{width=\linewidth}
    \begin{tabular}{llrrrrr}
    \toprule
    Task & Model & $\beta_0$ & $a$ & $b$ & $c$ & $R^2$ \\
    \midrule
    CP  & 8B  & -0.020 & -0.602 & -0.496 & -0.027 & 0.811 \\
    CP  & 14B & -0.561 & -0.442 & -0.373 &  0.007 & 0.785 \\
    MMD & 8B  & -0.590 & -0.208 & -0.290 & -0.106 & 0.916 \\
    MMD & 14B & -0.641 & -0.342 & -0.238 & -0.057 & 0.869 \\
    HT  & 8B  &  0.592 & -0.363 & -0.168 & -0.012 & 0.853 \\
    CP  & 1.7B & -0.498 & -0.076 & -0.088 &  0.001 & 0.724 \\
    CP  & 4B   & -0.524 & -0.190 & -0.157 & -0.017 & 0.870 \\
    MMD & 4B   & -0.223 & -0.418 & -0.248 &  0.024 & 0.732 \\
    \bottomrule
    \end{tabular}
    \end{adjustbox}
\end{table}

The depth--breadth split determines where on the fitness landscape a 
run lands. Here we characterize the landscape itself.

\paragraph{The fitness landscape.}
For a practitioner with fixed budget $C$, choosing $T$ is equivalent
to selecting a point on the fitness surface along the budget slice of
$C$. As \Cref{fig:surface_all8} shows, the effective-FLOPs surface
(top) is approximately linear in $T$ and remains weakly dependent on $T$ at fixed $C$, so fitness differences along the depth axis reflect allocation choices rather than hidden cost variation. The fitness surface (bottom) reveals different geometries across tasks: CP shows a broad plateau where many depth allocations achieve near-equivalent fitness, whereas MMD shows an interior ridge where only a balanced depth allocation achieves the highest in-sweep fitness. The same protocol and budget range thus produce qualitatively different optimization landscapes, raising the question of what underlying task properties drive the difference.

\noindent\textbf{An empirical bilinear form.}~
We fit a parametric model to the sub-ceiling cells ($V<0.97$), with log fitness gap as the response and depth and breadth as predictors:
\begin{equation}\label{eq:bi-law}
    \log(1-V) = \beta_0 + a\log T + b\log N + c\log T\log N.
\end{equation}
A budget-only model ($c{=}0$, $a{=}b$) reaches
$R^2\!\approx\!0.74$--$0.78$; the bilinear form reaches
$R^2\in[0.75,0.92]$ across all three tasks
(\Cref{tab:production_law}). Allocation, with budget, carries
the signal. The coefficient $c$ characterizes the regime (interior at large negative $|c|$, corner near $|c|\approx 0$; full boundary analysis in \Cref{app:boundary})
in a way prior inference-time scaling work reports descriptively but
does not pin down~\cite{snell2024scaling, inoue2025abmcts}; higher-order
terms ($\log^{2}T$, $\log^{2}N$) do not materially improve the fit
(\Cref{app:production_law}).


\begin{figure*}[t]
    \centering
    \includegraphics[width=0.9\linewidth]{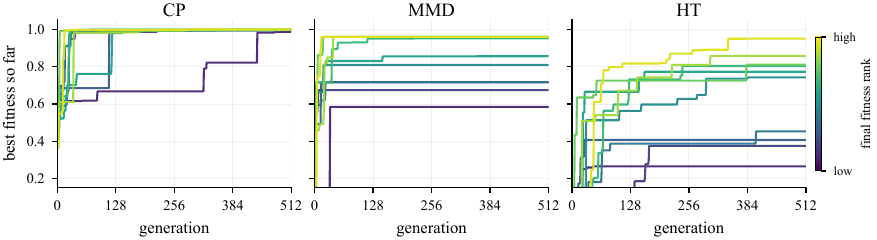}
    \caption{Per-run greedy fitness trajectories at $T{=}512$, $N{=}1$ on Qwen3-8B. Greedy reliably reaches high fitness in CP, while several seeds in MMD and HT stagnate at suboptimal values.}
    \label{fig:greedy_per_seed_trajectory}
\end{figure*}

Among the four coefficients, $c$ alone controls geometry: small $|c|$ leaves the budget slice near-separable with the optimum at the all-depth corner, while large negative $|c|$ bends it inward to a balanced interior optimum, with plateau half-width $\propto 1/\sqrt{|c|}$ (\Cref{app:plateau_width}).
MMD sits in the latter limit on both
capable models; CP and HT sit in the former
(\Cref{tab:production_law}, with the near-zero rows discussed in
\Cref{app:boundary}). The case studies (\Cref{app:case_study}) are consistent with a mechanism we call asymmetric proposal mass: each task
admits one high-fitness algorithmic family, but the LLM's base rate
for this family appears to differ across tasks, so where the good family is rare,
breadth raises the probability that at least one parallel trajectory
anchors on it before depth can refine within it.



\section{Modeling Evolving Trajectories through Multi-Armed Bandits}
\label{sec:MAB_for_trajectory_allocation}

\Cref{sec:scaling_law} characterized within-run depth-breadth allocation on capable model and task. However, the self-evolving process itself remains highly stochastic. Specifically, we run the experiments 10 times under the same allocation configuration (sequential: $T=512, N=1$) and analyze each \emph{greedy} run's performance by tracing its fitness score trajectory throughout the game in \Cref{fig:greedy_per_seed_trajectory}. Evolving may dramatically fail from time to time by converging at a low score without further improvement even under the \emph{same configuration}, producing a score distribution (\Cref{app:distribution}) rather than a single value (\Cref{app:case_study}), a spread that within-run allocation cannot remove. A second allocation lever is therefore needed, one that operates between trajectories rather than within them.

\subsection{Bandits Adaptation}
MAB provides a framework for sequential decision-making in uncertain environments. In the standard setting, the player chooses an action to take from a finite action set $[K]$ at each round $c \in [C] := \{ 1, 2, \dots, C \}$ and observes the associated reward, which may immediately factor into the decision in the next round. Modeling cross-trajectory allocation through this framework, we propose Bandits-based Self-Evolving (\texttt{BaSE}) in \Cref{alg:base}.

\begin{algorithm}[t]
\caption{Bandits-based Self-Evolving (\texttt{BaSE})}
\label{alg:base}
\footnotesize
\begin{algorithmic}[1]
\Require Task $q$, base LLM $f$, prompt generator $h$, budget (number of LLM calls) $C$, number of runs $K$

\State Initialize $c_i \gets 1$ for all $i \in [K]$

\ForAll{$i \in [K]$ \textbf{in parallel}}
    \State $\mathrm{prompt} \gets h \left( q \right)$
    \State $\mathrm{response}_i^{(c_i)} \gets f(\mathrm{prompt})$
    \State $\mathrm{score}_i \gets \mathrm{Eval}\left( q, \mathrm{response}_i^{(c_i)} \right)$
\EndFor

\For{$c = K+1, \dots, C$}
    \State $i \gets \mathrm{MAB}\left( 1, \dots, K \right)$ 
    \State \label{line:base-parent-sampling} $\mathrm{prompt} \gets h \left( q, \left( \mathrm{response}_i^{(j)}, \mathrm{score}_i^{(j)} \right)_{j \in [c_i]} \right)$
    \State $\mathrm{response}_{i}^{(c_{i}+1)} \gets f(\mathrm{prompt})$
    \State $\mathrm{score}_{i}^{(c_i + 1)} \gets \mathrm{Eval} \left( q, \mathrm{response}_{i}^{(c_{i} + 1)} \right)$
    \State $c_{i} \gets c_{i}+1$
\EndFor

\State $(i^\star, j^\star) \gets \arg\max_{i \in [K],\, j \in [c_i]} \mathrm{score}_i^{(j)}$
\State \Return $\mathrm{response}_{i^\star}^{(j^\star)}$

\end{algorithmic}
\end{algorithm}

Specifically, we set the action set to $K$-runs with different latent trajectories; rewards to fitness scores from the evaluator; and each round to a step of the self-evolving process. Instead of committing all budget to one evolving trajectory as in \Cref{alg:self_evolving}, \Cref{alg:base} maintains $K$ parallel runs and adaptively decides which trajectory should receive the next round of calls based on runs' historical rewards (historical fitness scores) and number of calls through bandit policies. In other words, the algorithm splits the fixed budget across multiple trajectories and adaptively redistributes subsequent calls toward the optimal trajectory with the highest fitness score. Intuitively, the allocator works by deciding whether to exploit its observed potential (pull a trajectory that already appears promising), or to explore alternatives (pull a less-tested trajectory whose future potential remains uncertain). In the experiments, we use $K=10$ by default, i.e., 10 runs (trajectories) in parallel.

\newcommand{\ms}[2]{\makecell[c]{#1\\[-1pt]{\scriptsize $\pm$\,#2}}}
\newcommand{\bms}[2]{\makecell[c]{\textbf{#1}\\[-1pt]{\scriptsize $\pm$\,#2}}}

\begin{table*}[t]
\centering
\caption{Fitness scores (mean $\pm$ bootstrap standard error (SE)) across models, tasks, and evolution strategies at 512 LLM calls ($C{=}512$). The maximum fitness score in each (model, task) is \textbf{bolded}.}
\label{tab:fitness_scores}
\vspace{2pt}
\small
\renewcommand{\arraystretch}{1.10}
\setlength{\tabcolsep}{2.2pt}
\setlength{\arrayrulewidth}{0.4pt}

\begin{tabular*}{\textwidth}{@{\extracolsep{\fill}}llcccc|cc|ccc@{}}
\toprule
\multirow{2}{*}{\textbf{Model}} & \multirow{2}{*}{\textbf{Task}}
& \multicolumn{4}{c}{\textbf{Baselines}} & \multicolumn{5}{c}{\textbf{\texttt{BaSE}}} \\
\cmidrule(lr){3-6}\cmidrule(lr){7-11}
& & {\scriptsize\textbf{Greedy}} & {\scriptsize\textbf{OpenEvolve}} & {\scriptsize\textbf{CodeEvolve}} & {\scriptsize\textbf{ShinkaEvolve}}
& {\scriptsize\textbf{Random}} & {\scriptsize\textbf{Round-Robin}} & {\scriptsize\textbf{UCB}} & {\scriptsize\textbf{EXP3.P}} & {\scriptsize\textbf{Thompson}} \\
\specialrule{\lightrulewidth}{2pt}{2pt}

\multirow{3}{*}{\textbf{Qwen3-8B}}
& CP  & \ms{0.9985}{0.0005} & \ms{0.8279}{0.0145} & \ms{0.7692}{0.0202} & \ms{0.9986}{0.0003}
& \ms{0.9962}{0.0058} & \ms{0.9965}{0.0119} & \ms{0.9965}{0.0102} & \ms{0.9993}{0.0059} & \bms{1.0003}{0.0174} \\
& MMD & \ms{0.9597}{0.0039} & \ms{0.9561}{0.0143} & \ms{0.8414}{0.0689} & \ms{0.9099}{0.0240}
& \bms{0.9696}{0.0116} & \ms{0.9603}{0.0069} & \ms{0.9603}{0.0053} & \ms{0.9658}{0.0102} & \ms{0.9603}{0.0035} \\
& HT  & \ms{0.6780}{0.0718} & \ms{0.6061}{0.0768} & \ms{0.5168}{0.0627} & \ms{0.7379}{0.0373}
& \ms{0.7153}{0.0497} & \ms{0.7153}{0.0468} & \ms{0.8530}{0.0963} & \ms{0.8653}{0.1068} & \bms{0.8736}{0.1737} \\
\specialrule{\lightrulewidth}{2pt}{2pt}

\multirow{3}{*}{\textbf{Llama}}
& CP  & \ms{0.8432}{0.0496} & \ms{0.7033}{0.0657} & \ms{0.4568}{0.0330} & \ms{0.8305}{0.0362}
& \ms{0.9327}{0.0622} & \ms{0.9310}{0.0798} & \ms{0.9451}{0.0633} & \ms{0.9484}{0.0716} & \bms{0.9533}{0.0706} \\
& MMD & \ms{0.8424}{0.0215} & \ms{0.6378}{0.1024} & \ms{0.2315}{0.0847} & \ms{0.7253}{0.0453}
& \ms{0.9044}{0.1084} & \bms{0.9512}{0.1066} & \bms{0.9512}{0.1020} & \ms{0.8516}{0.1072} & \ms{0.8634}{0.0955} \\
& HT  & \ms{0.2538}{0.0762} & \ms{0.0000}{0.0000} & \ms{0.0000}{0.0000} & \ms{0.1512}{0.0437}
& \ms{0.1676}{0.1154} & \ms{0.2016}{0.0944} & \ms{0.2561}{0.1196} & \ms{0.2729}{0.1851} & \bms{0.4387}{0.2134} \\
\specialrule{\lightrulewidth}{2pt}{2pt}

\multirow{2}{*}{\textbf{Qwen3-14B}}
& CP  & \ms{0.9802}{0.0060} & \ms{0.8577}{0.0228} & \ms{0.8768}{0.0134} & \ms{0.9981}{0.0003}
& \ms{0.9955}{0.0087} & \ms{0.9972}{0.0074} & \ms{0.9972}{0.0074} & \ms{0.9967}{0.0114} & \bms{1.0003}{0.0143} \\
& MMD & \ms{0.9816}{0.0059} & \ms{0.9949}{0.0012} & \ms{0.9934}{0.0021} & \ms{0.9924}{0.0037}
& \ms{0.9963}{0.0324} & \ms{0.9958}{0.0351} & \ms{0.9969}{0.0356} & \ms{0.9922}{0.0283} & \bms{0.9983}{0.0163} \\
\bottomrule
\end{tabular*}
\end{table*}

\begin{table*}[t]
    \centering
    \caption{Minimum generation (Gen.) and cumulative FLOPs ($\times 10^{15}$) required for $\geq 90\%$ samples to reach thresholds ($\geq \tau$) with Qwen3-8B as the base LLM. Unreached thresholds are denoted as ``---''. The minimum generation and FLOPs for each threshold are \textbf{bolded}.}
    \label{tab:threshold_iteration_flops}
    \begin{adjustbox}{width=\linewidth,center}
    {
    \begin{tabular}{@{}cc *{16}{c}@{}}
    \toprule
    \textbf{Task} & $\boldsymbol{\tau}$ 
    & \multicolumn{8}{c}{\textbf{Baselines}}
    & \multicolumn{8}{c}{\textbf{\texttt{BaSE}}} \\
    \cmidrule(lr){3-10}\cmidrule(lr){11-18}
    & 
    & \multicolumn{2}{c}{Greedy} 
    & \multicolumn{2}{c}{OpenEvolve} 
    & \multicolumn{2}{c}{CodeEvolve} 
    & \multicolumn{2}{c}{ShinkaEvolve} 
    & \multicolumn{2}{c}{Random} 
    & \multicolumn{2}{c}{UCB} 
    & \multicolumn{2}{c}{EXP3.P} 
    & \multicolumn{2}{c}{Thompson} \\
    \cmidrule(lr){3-4}\cmidrule(lr){5-6}\cmidrule(lr){7-8}\cmidrule(lr){9-10}
    \cmidrule(lr){11-12}\cmidrule(lr){13-14}\cmidrule(lr){15-16}\cmidrule(lr){17-18}
    & & Gen. & FLOPs & Gen. & FLOPs & Gen. & FLOPs & Gen. & FLOPs 
      & Gen. & FLOPs & Gen. & FLOPs & Gen. & FLOPs & Gen. & FLOPs \\
    \midrule

    \multirow{3}{*}{CP} 
    & 0.95 & 152 & 182.12 & --- & --- & --- & --- & 45 & 57.90 & 56 & 66.58 & \textbf{16} & \textbf{18.19} & 139 & 173.57 & \textbf{16} & \textbf{18.19} \\
    & 0.99 & 359 & 411.93 & --- & --- & --- & --- & \textbf{73} & \textbf{90.15} & 138 & 170.88 & 85 & 100.79 & 207 & 256.12 & 85 & 101.22 \\
    & 0.999 & --- & --- & --- & --- & --- & --- & --- & --- & --- & --- & --- & --- & --- & --- & \textbf{327} & \textbf{395.47} \\
    
    \midrule

    \multirow{3}{*}{MMD} 
    & 0.80 & 212 & 318.46 & 119 & 181.48 & --- & --- & 71 & 112.00 & 26 & 32.34 & \textbf{8} & \textbf{12.41} & 40 & 52.72 & \textbf{8} & \textbf{12.41} \\
    & 0.90 & 296 & 436.55 & --- & --- & --- & --- & --- & --- & 119 & 179.07 & \textbf{92} & \textbf{117.63} & 95 & 142.59 & 105 & 144.75 \\
    & 0.95 & 485 & 656.81 & --- & --- & --- & --- & --- & --- & 132 & 199.95 & \textbf{92} & \textbf{117.63} & 114 & 172.74 & 106 & 146.75 \\

    \midrule

    \multirow{2}{*}{HT} 
    & 0.50 & --- & --- & --- & --- & --- & --- & 201 & 189.28 & 92 & 99.90 & \textbf{60} & \textbf{66.90} & 102 & 108.67 & 74 & 79.45 \\
    & 0.70 & --- & --- & --- & --- & --- & --- & --- & --- & 367 & 381.23 & 125 & 130.55 & 209 & 213.22 & \textbf{101} & \textbf{104.53} \\

    \bottomrule
    \end{tabular}
    }
    \end{adjustbox}
\end{table*}

\subsection{Our Results}
\noindent\textbf{Fitness Score.}~ 
As shown in \Cref{tab:fitness_scores}, we report the average fitness score across three tasks, model families, and self-evolving strategies at $C=512$ LLM calls. \texttt{BaSE} achieves the best or near-best fitness across models and tasks. For example, Thompson reaches $0.8736$ on Qwen3-8B HT, clearly above greedy ($0.6780$) and ShinkaEvolve ($0.7379$); on Llama HT, it obtains $0.4387$, outperforming all baselines by a large margin; and on Qwen3-14B CP, it attains the highest average fitness score of $1.0003$. 

The improvement reflects the value of allocating the fixed budget across multiple evolving trajectories, rather than modifying the refinement dynamics within a trajectory, while keeping the total LLM-call budget fixed. Indeed, under the same fixed computation (generation) budget, trajectory diversification itself helps avoid committing the full budget to a low-potential run but adaptively allocates budget to more promising trajectories.

Although BaSE attains the strongest mean across all reported model-task pairs,  we note that the magnitude of the improvement varies with task difficulty and available headroom. On saturated tasks such as CP, several baselines already approach the score ceiling; for example, Qwen3-8B greedy and ShinkaEvolve achieve $0.9985$ and $0.9986$, while BaSE-Thompson reaches $1.0003$ with CI $[0.9980, 1.0004]$. Hence, the CP gains are necessarily small. By contrast, BaSE shows clearer benefits on harder settings such as HT, where Qwen3-8B BaSE-Thompson reaches $0.8736$ compared with the best non-BaSE score of $0.7379$, and Llama BaSE-Thompson reaches $0.4387$ while all non-BaSE methods remain below $0.2538$.

\noindent\textbf{Cross-Trajectory Allocation with Parent Sampling Protocols.}~
\texttt{BaSE} is orthogonal to parent-sampling strategies: \texttt{BaSE} decides which trajectory receives the next LLM call, while parent sampling decides which historical response is used to refine a selected trajectory. Thus, \Cref{line:base-parent-sampling} in \Cref{alg:base} can replace the vanilla prompt generator with existing parent-sampling methods. In \Cref{sec:pairwise_fitness}, we evaluate this combination by pairing \texttt{BaSE} with different parent-sampling baselines and comparing against greedy variants under different breadth-depth allocations. As shown in \Cref{tab:pairwise_fitness}, these strategies can be naturally integrated into \texttt{BaSE}, often improving their best fitness scores, especially in unstable settings with high trajectory variance.

\noindent\textbf{Sample-Efficient Threshold Reaching.}~
Beyond final fitness, we evaluate how efficiently each method reaches a target fitness threshold $\tau$ under a fixed budget. Following the time-to-threshold metric of \citet{JMLR:v8:taylor07a}, we report the earliest generation $G$ and corresponding cumulative FLOPs at which $90\%$ of bootstrap samples reach $\tau$. As shown in \Cref{tab:threshold_iteration_flops}, \texttt{BaSE} reaches most thresholds with fewer generations and FLOPs, especially on MMD and HT. For example, on MMD, UCB reaches $\tau=0.95$ at $92$ generations, while greedy requires $485$; on HT, UCB reaches $\tau=0.50$ within $60$ generations, whereas greedy, OpenEvolve, and CodeEvolve never reach it. Compared with random allocation, Thompson reaches the same thresholds with $\sim$40\% fewer generations on average across the seven reached cells. Full results across three models are deferred to \Cref{sec:full_threshold_comparison}.


\noindent\textbf{Ablation on Arm-Pool Size.}~
We ablate the number of parallel runs $K$ (number of arms) used by \texttt{BaSE} and defer the full results to \Cref{sec:bandit_ablation_arms} (\Cref{tab:bandit_ablation_arms}). The results show that moderate arm pools usually provide the best trade-off between trajectory diversity and per-run refinement depth. In particular, $K \in \{5,10,20\}$ often achieves the strongest performance, while very small pools can lack sufficient exploration and overly large pools such as $K=50$ may dilute the refinement budget across too many shallow trajectories. 



\section{Discussion}

\paragraph{Prompting and allocation move different quantities.}
The CP results in \Cref{tab:fitness_scores} appear to make ShinkaEvolve a rather strong baseline ($0.9986$ on Qwen3-8B), while inspection of the prompts in \Cref{app:prompts} shows this advantage could be prompt-induced: ShinkaEvolve's CP system prompt explicitly suggests \texttt{scipy.optimize}. In contrast, our method uses exactly the same prompt as OpenEvolve, steering away from quick-win hints. Under such a constraint, our greedy baseline reaches $0.9985$ on the same task, which is a $+0.17$ gap over OpenEvolve. As \Cref{app:case_study} documents, greedy closes on ShinkaEvolve because the LLM eventually discovers \texttt{scipy} despite the fact that it is not on top of a privileged prompt. The improvements we report are therefore achieved through allocation rather than through prompt tuning. The prompt hint affects \emph{when} a promising algorithmic family appears, whereas allocation determines how effectively promising trajectories are subsequently refined. Therefore, both prompt engineering and allocation are complementary levers.



\paragraph{Model capability and prompt set the ceiling that allocation can reach.}
Allocation amplifies an existing signal but cannot create one. On 
Llama HT (\Cref{tab:fitness_scores}), OpenEvolve and CodeEvolve collapse 
to $0.0$ (no run produces a valid configuration in 512 calls)
and even Thompson only recovers to $0.4387$, an order of 
magnitude below Qwen3-8B on the same task. \Cref{app:effect_model_size} 
shows the same pattern across the full sweep: when a model fails 
to cross the capability threshold on a relatively difficult task, depth gains are statistically indistinguishable from selection noise. The prompt sets a separate 
ceiling: ShinkaEvolve's HT prompt (\Cref{app:prompts}) contains a 
``CRITICAL --- degeneracy warning'' about collinear-triplet failures 
that no other baseline carries, and this directly addresses the 
failure mode diagnosed in \Cref{app:case_study} (HT run~5). Capability 
and prompt thus jointly determine the achievable ceiling; allocation 
governs how efficiently a method approaches it.

\paragraph{\texttt{BaSE} is complementary to within-run mechanisms.}
\texttt{BaSE} operates at trajectory granularity, while the parent-sampling 
protocols in *Evolve systems operate within a single run; the two 
compose. The improvement of \texttt{BaSE} over a single-trajectory baseline 
admits a clean two-step decomposition: \emph{Greedy/Island} 
$\to$ \emph{Random} isolates the pool effect (drawing $K$ 
independent trajectories and returning the best), and 
\emph{Random} $\to$ \texttt{BaSE} isolates the allocation effect 
(adaptively routing compute toward promising trajectories). The 
pool effect is licensed by the per-run heterogeneity documented 
in \Cref{fig:greedy_per_seed_trajectory} and 
\Cref{app:case_study}. The allocation effect is then 
isolated by the Random vs.\ \texttt{BaSE} gap in 
\Cref{tab:threshold_iteration_flops}. Both effects survive when \texttt{BaSE} replaces the prompt generator of OpenEvolve, CodeEvolve, or ShinkaEvolve 
(\Cref{tab:pairwise_fitness} in \Cref{sec:pairwise_fitness}), with 
the largest gains where the underlying protocol leaves trajectory 
variance unexploited and the smallest where it has already 
saturated the task.

\section{Conclusion}

In conclusion, we presented the first fixed-budget characterization of depth--breadth allocation in LLM-guided evolutionary search, revealing structured, task-dependent fitness surfaces and an empirical bilinear regularity. We then showed that fixed allocations still leave substantial cross-trajectory heterogeneity. Motivated by this observation, we introduced \texttt{BaSE}, a bandit-based allocator that improves expected fitness and threshold-reaching efficiency over strong \emph{*Evolve} baselines without changing the base model, evaluator, or prompt design. This highlights compute allocation as a key lever for reliable finite-budget search.

\section*{Limitations}
Our main experiments focus on three geometric optimization-based coding problems (Circle Packing, MinMaxDist, Heilbronn Triangle) inherited from the OpenEvolve suite. \Cref{app:lpo} provides additional evidence that the scaling regularities extend to LLM Prompt Optimization (LPO), but broader cross-domain generalization is not fully validated. Secondly, our model coverage spans one open-weight family (Qwen3 1.7B--14B) plus Llama-3.1-8B; closed-source frontier models on which AlphaEvolve and FunSearch were originally evaluated are not tested. Thirdly, the bilinear fit is a descriptive summary, not a predictive law: the interaction coefficient is task-specific and not statistically significant on Circle Packing, and whether the form transfers to unseen tasks is left open. Finally, the capability gate is treated as a per-cell binary; a continuous characterization across model capability and a low-cost diagnostic probe are left to future work. For \texttt{BaSE}, the bandit policies serve as a heuristic implementation for balancing exploration and exploitation in cross-trajectory compute allocation with validated empirical benefits. The theoretical regret bounds and optimality analysis remain an open problem. 


\section*{Acknowledgments}

This work was supported in part by institutional research support associated with the Melchor Visiting Assistant Professorship at the University of Notre Dame and by computational resources provided through ACCESS allocation CIS260386. This research was also supported by the National Artificial Intelligence Research Resource (NAIRR) Pilot and the Anvil supercomputer supported by the National Science Foundation (award NSF-OAC 2005632) at Purdue University.
Tianfan Fu is supported by the Young Scientists Fund (C Class) of the National Natural Science Foundation of China (Grant No.~62506154), the Fundamental Research Funds for the Central Universities, the Nanjing University International Collaboration Initiative (Grant No.~020214380129), and the ``111 Center'' (No.~B26023).

\bibliography{custom}

\appendix

\newpage

\addtocontents{toc}{\protect\setcounter{tocdepth}{2}}
\renewcommand{\contentsname}{Appendix Contents}
\tableofcontents

\newpage

\section{Effective FLOPs}
\label{app:flops}

The protocols differ in cache behavior as a structural consequence of their prompt construction, where greedy reuses a single parent prefix across batched siblings in a generation, while island rewrites the parent + inspirations per call. 

\paragraph{Formula.}
We charge each completed LLM call
\[
\textsc{FLOPs}_{\text{call}} = 2\,P_{\text{active}}\,(p_{\text{uncached}} + p_{\text{out}}),
\]
where $P_{\text{active}}$ is the active parameter count, 
$p_{\text{uncached}} = p_{\text{prompt}} - p_{\text{cached}}$, and 
$p_{\text{out}}$ is the completion token count. Per-run FLOPs sums over 
all attempts, including retries. Token counts and cache statistics are 
read directly from vLLM's per-request \texttt{usage} records. The 
factor $2\,P_{\text{active}}$ per token is the forward-pass cost in the 
$C \approx 6ND$ in the spirit of Chinchilla~\cite{hoffmann2022chinchilla}, using only the forward $2N$ component as no backward pass occurs in LLM-guided 
evolution. As in Chinchilla, this convention omits sub-leading terms 
(attention score computation, softmax, layer norms, embeddings, 
sampling); Chinchilla's reports agreement with detailed 
accounting to within 10\% over two orders of magnitude.

\paragraph{Active parameters.}
All models in our sweep (Qwen3 1.7B/4B/8B/14B and Llama-3.1-8B) are 
dense, so $P_{\text{active}}$ equals the total parameter count. The 
notation is retained for compatibility with MoE models.

\paragraph{Prefix-cached tokens.}
Cached prompt tokens are subtracted because vLLM's PagedAttention 
reuses their KV states without recomputing the corresponding forward 
pass. Cache hit rates in our sweeps range from 94\% to 98\%; omitting 
this subtraction over-reports FLOPs by 6--9\%. The residual attention 
computation over cached tokens is sub-leading and is omitted, 
consistent with the convention above.


\paragraph{Non-embedding variant.}
Some scaling-law conventions use non-embedding parameter counts in 
place of $P_{\text{active}}$. Substituting would multiply every reported 
FLOPs value by a fixed per-model factor 
$P_{\text{non-embed}}/P_{\text{active}}$, preserving all cross-model 
ratios and orderings used in our analysis.

\section{Task Details}
\label{app:task_details}

For all three tasks, raw objectives are normalized by the best
published construction so that $\textsc{fitness}{=}1.0$ matches
the state of the art:

\textbf{Circle Packing (CP, $n{=}26$).} Pack $26$ non-overlapping circles into the unit
square to maximize the sum of their radii. Fitness is normalized to the AlphaEvolve
reference of $2.635$ for $n=26$, so normalizer: $\textsc{fitness} = \textsc{sum}_{\text{radii}} / 2.635$.

\noindent\textbf{MinMaxDist (MMD, $n{=}16, d{=}2$).}
Place $16$ points in the plane to maximize
the ratio of minimum to maximum pairwise distance, $d_{\min} / d_{\max}$. Fitness is the
squared ratio normalized to the benchmark of
$1/\sqrt{12.889266112} \approx 0.2786$, normalizer:
$\textsc{fitness} = (d_{\min}/d_{\max})^{2} \cdot 12.889266112$; The configuration space contains a sharp deterministic attractor at fitness $\approx 0.9603$ corresponding to a $5{+}11$
construction.

\textbf{Heilbronn Triangle (HT, $n{=}11$).}
Place $11$ points on or inside the equilateral triangle with vertices
$(0,0)$, $(1,0)$, and $(0.5,\sqrt{3}/2)$. The evaluator enumerates all
$\binom{11}{3}$ triplets and computes the smallest triangle area,
normalized by the area of the containing equilateral triangle. The final
score divides this normalized minimum area by the reference
$r_{\mathrm{AE}}=0.036529889880030156$.

Note that subsequent *Evolve runs have marginally surpassed the $2.635$ AlphaEvolve CP reference, so $\textsc{fitness}$ values slightly above $1.0$  reflect this margin rather than a novel construction; we retain $2.635$ as the normalizer for harness consistency with prior work.

\section{The Capability Gate of the Collapse}
\label{app:effect_model_size}

The fitness--FLOPs envelope of \Cref{sec:envelope} collapses across
model size only partially: capable models lie on one curve, weaker
ones fall below it. To characterise the boundary we compute, per
model$\times$task cell, the best-of-$N$ fitness $V_{T{=}1}$, the best
mean fitness over all tested depths $V_{\max}$, the depth gain
$\mathrm{penBoN}=V_{\max}-V_{T{=}1}$, and a permutation $p$-value for
that gain (\Cref{tab:effect_model_size}).

Under the null that depth has no effect, we shuffle per-seed fitness
across~$T$, recompute $\mathrm{penBoN}$ on each shuffle, and let $p$
be the fraction of $20{,}000$ shuffles meeting or exceeding the
observed value. A bootstrap CI is uninformative because $V_{\max}$
already takes a max over $\sim\!10$ depth cells, forcing
$\mathrm{penBoN}\ge 0$ on every resample; the permutation null
re-maxes too and so absorbs that selection inflation.

\begin{table}[t]
\centering\small
\caption{Capability gate per model$\times$task cell. $V_{T{=}1}$ is the
best-of-$N$ fitness, $V_{\max}$ the best over all depths,
$\mathrm{penBoN}=V_{\max}-V_{T{=}1}$, and $p$ the depth permutation test.
1.7B/4B evaluated at $C{=}128$, all others at $C{=}512$.}
\label{tab:effect_model_size}
\begin{adjustbox}{width=1\linewidth,center}
\begin{tabular}{llrrrr}
\toprule
Model & Task & $V_{T{=}1}$ & $V_{\max}$ & penBoN & $p$ \\
\midrule
1.7B  & CP & 0.582 & 0.615 & $+0.033$ & 0.48 \\
4B    & CP & 0.738 & 0.790 & $+0.052$ & 0.30 \\
8B    & CP & 0.989 & 0.999 & $+0.009$ & 0.13 \\
14B   & CP & 0.926 & 0.980 & $+0.054$ & 0.07 \\
Llama & CP & 0.843 & 0.843 & $+0.000$ & 1.00 \\
\midrule
1.7B  & MMD & 0.716 & 0.748 & $+0.031$ & 0.16 \\
4B    & MMD & 0.694 & 0.906 & $+0.213$ & 0.002 \\
8B    & MMD & 0.944 & 0.960 & $+0.016$ & 0.64 \\
14B   & MMD & 0.867 & 0.981 & $+0.115$ & ${<}10^{-3}$ \\
Llama & MMD & 0.673 & 0.843 & $+0.170$ & 0.23 \\
\midrule
8B    & HT & 0.339 & 0.672 & $+0.333$ & 0.002 \\
Llama & HT & 0.047 & 0.253 & $+0.206$ & 0.013 \\
\bottomrule
\end{tabular}
\end{adjustbox}
\end{table}

The depth signal turns on with scale within Qwen3. At 1.7B
$\mathrm{penBoN}$ is small and not significant on either task
($p\ge 0.16$). At 4B it appears on MMD ($+0.213$, $p=0.002$) but not
on CP ($p=0.30$). On CP and MMD, 8B and 14B reach $V_{\max}\ge 0.96$
on every tested cell, near the task ceiling, and depth gains compress
to small residuals, while the exception is 14B/MMD, $+0.115$ at
$p<10^{-3}$. On the unsaturated HT task, 8B/HT instead shows the
largest clean depth gain in the table: $V$ rises from $0.339$ to
$0.672$ ($+0.333$, $p=0.002$). The depth signal therefore turns on
whenever a capable model has room above best-of-$N$. The threshold is
per (model, task), not per model: 4B sits on either side depending
on the task.

Most striking is Llama-3.1-8B. On CP best-of-$N$ already reaches
$0.843$ and depth does not improve it ($\mathrm{penBoN}=0$,
$p=1.00$). On MMD the nominal $+0.170$ depth gain is consistent with
max-selection noise ($p=0.23$). On HT depth does compound ($V$
rises from $0.047$ to $0.253$ at $p=0.013$), but the absolute level
stays at $0.25$, far below the $0.67$ that Qwen3-8B reaches on the
same task. Despite its 8B parameters, Llama-3.1-8B never achieves
both a significant depth gain and a non-trivial absolute level on
any tested task. Crossing the threshold is thus not a matter of
parameter count.

\section{Case Studies: Why Runs Stagnate}
\label{app:case_study}

The main text (\Cref{fig:greedy_per_seed_trajectory}) shows that ten
greedy runs of an identical configuration produce very different
final fitness: some climb smoothly, others stop improving early and
sit at a low score for the rest of the $C{=}512$ budget. This
appendix opens up that variability and asks, at the level of the
actual Python code the LLM produced, \emph{why} certain runs get
stuck.

Each run is a chain of accepted programs. At every generation,
greedy keeps the single best-scoring program seen so far and asks
the LLM to mutate it. A child is accepted only if it strictly beats
this running best. So a run gets ``stuck'' when, for many
generations in a row, the LLM proposes mutations that the evaluator
either rejects (returns a worse score) or cannot score at all
(invalid output). To understand a stuck run we therefore have
to look at two things: (i) what program the run is anchored on, and
(ii) why no proposed mutation beats it.

\begin{figure*}[t]
    \centering
    \includegraphics[width=0.95\linewidth]{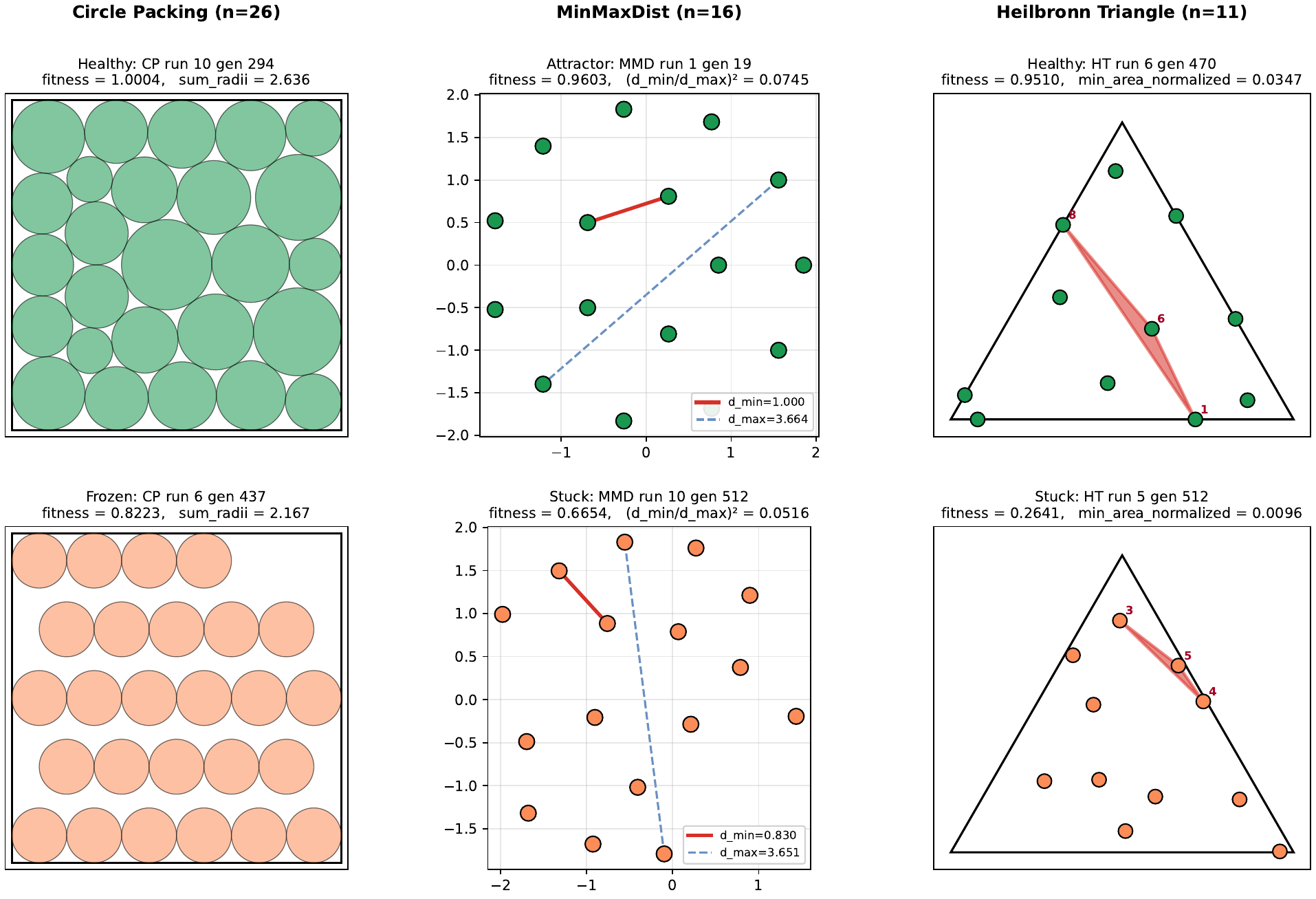}
    \caption{One healthy (top) and one stuck (bottom) greedy run per
    task at Greedy $T{=}512$, $N{=}1$ on Qwen3-8B. Each title gives the run
    number (out of 10), the generation at which the displayed
    program was the running best, the normalized fitness, and the
    task's raw metric. \textbf{CP}: the unit square is filled with
    26 non-overlapping circles; \textbf{MMD}: 16 points; the red bar is the closest pair
    ($d_{\min}$) and the blue dashed bar the farthest pair
    ($d_{\max}$); \textbf{HT}: 11 points inside the equilateral
    triangle; the red shaded triplet is the smallest of the
    $\binom{11}{3}{=}165$ triangles, which sets the score.}
    \label{fig:case_study}
\end{figure*}

\Cref{fig:case_study} shows one healthy and one stuck run per task,
using the actual point/circle configurations the anchored program
outputs. The rest of this appendix walks through each task in turn.

\subsection{Circle Packing: \texttt{scipy} acquisition is the
gating event}
\label{app:case_study_cp}

\paragraph{The task in one sentence.} Pack $26$ non-overlapping
circles into the unit square so that the sum of their radii is as
large as possible; the AlphaEvolve reference is
$\sum_i r_i = 2.635$, according $\text{fitness}{=}1.0$ base on this sum of radii matches the state of the art.

\paragraph{Two regimes the LLM oscillates between.} On CP the LLM
writes programs that fall into two clearly different families:
\begin{enumerate}
    \item[(a)] \emph{Hand-coded ring constructors:} the program lays
    $26$ circles down in geometric rows (e.g. $6{+}5{+}6{+}5{+}4$
    hexagonal rows) and assigns each circle the same fixed radius
    based on the row spacing. No optimization is performed. The
    best score this family can produce in our runs is
    $\sim\!0.89$, because uniform radii waste space in the corners
    and at the row endings.
    \item[(b)] \texttt{scipy.optimize.minimize}-based programs: the
    same row layout is used as an initial guess, but the program
    then calls SLSQP or COBYLA with non-overlap and box-containment
    constraints, letting both centers and radii float. These
    programs reach $\sim\!1.0$.
\end{enumerate}

Acquiring \texttt{scipy} is therefore the gating event for high
fitness on CP: every run that finishes above $0.99$ is anchored on
a \texttt{scipy} program; the one run that does not is the one
where the LLM took longest to write a \texttt{scipy} call.

\paragraph{Healthy example --- run 10.} At generation 13, the LLM
proposes the first child that imports \texttt{scipy} and wraps the
ring layout in a constrained SLSQP call. The accepted score jumps
from $0.5584$ to $0.8396$ in a single generation, a $+0.28$
improvement that no later generation matches. Within another three
\texttt{scipy}-family accepts the run is at $0.9981$ (gen $24$);
few more refinements over the next generations push it to $\sum_i r_i = 2.636$ (\Cref{fig:case_study}).

\paragraph{Stuck example --- run 6.} The LLM proposes hand-coded
ring constructors for the first $437$ generations of this run.
Within that family, the anchored program slowly improves through
$5$ accepts from $0.36$ (initial) to $0.67$ (gen $88$), then sits
at $0.67$ for $233$ generations, then jumps to $0.79$ (gen $321$)
and $0.82$ (gen $327$) on two more ring-rewrites with non-hex
vertical spacing, and freezes at $0.82$ for another $110$
generations. The displayed gen-$437$ best
($\sum_i r_i = 2.167$, panel ``CP run 6 gen 437'' in
\Cref{fig:case_study}) is the entire visible cost of this stretch:
hundreds of attempts to improve a constructor whose intrinsic
ceiling is below $0.89$. The first \texttt{scipy} child finally arrives at generation $438$, producing a $+0.16$ jump in one step (from $0.822$ to $0.983$).

\subsection{MinMaxDist: the analytic attractor wins; \texttt{scipy}
\emph{poisons} the run}
\label{app:case_study_mmd}

\paragraph{The task in one sentence.} Place $16$ points in the
plane to maximize $(d_{\min}/d_{\max})^2$ --- the squared ratio of
the closest pair to the farthest pair --- normalized so that
$\text{fitness}{=}1.0$ matches the AlphaEvolve reference.

\paragraph{An analytic attractor exists.} For this problem, the
$5{+}11$ regular-polygon construction --- $5$ points uniformly on
an inner circle of radius $r$, $11$ points uniformly on an outer
circle of radius $R = r\,(1 + 2\sin(\pi/5))$ --- reaches the
mathematical optimum for this split at fitness $0.9603$. It
involves no optimizer at all: just \texttt{np.linspace} and
\texttt{np.cos}/\texttt{sin} calls in roughly fifteen lines of
code.

\paragraph{\texttt{scipy} is the \emph{bad} family on MMD.}
Counter-intuitively to the CP story, on MMD the LLM's
\texttt{scipy}-based programs are exactly the ones that get stuck.
Across the ten runs:
\begin{itemize}
    \item 3 Runs finish above the attractor ($0.9603$), all \texttt{scipy}-free: pure analytic $5{+}11$ polygons.
    \item 6 Runs finish below the attractor
    ($0.58$--$0.95$). All six final programs call either
    \texttt{differential\_evolution} or \texttt{minimize}, on an
    objective of the form $-(d_{\min}/d_{\max})^2$.
    \item Run 5 finishes at $0.86$ without \texttt{scipy} but also
    without the polygon construction.
\end{itemize}

The reason the \texttt{scipy} programs fail is that the objective
$-(d_{\min}/d_{\max})^2$ is highly non-convex; numerical optimizers
from a generic initial guess (4$\times$4 grid, 16-gon, etc.)
converge to local optima where one pair of points is much closer
than the rest, dragging $d_{\min}$ down. The first such program
greedy accepts becomes the run's permanent anchor, because future
proposed mutations have to beat its specific stochastic local
optimum in a single re-evaluation.

\paragraph{Healthy Attractor example --- run 1.} The run climbs in $4$
accepts from $0.02$ (initial: $16$ random Gaussian points) through intermediates ($0.72$, $0.86$, $0.89$) to
$0.9603$ at gen $19$, where the LLM writes the closed-form polygon
construction. The displayed gen-$19$ best (\Cref{fig:case_study}) shows two clean concentric rings
with all inner-to-outer distances equal; the closest pair sits
between adjacent outer points and the farthest pair spans a
diameter.

\paragraph{Stuck example --- run 10.} The run accepts five different scipy-based programs (gens 2, 3, 5, 7, 8), cycling through scipy.minimize from a 16-gon, then concentric squares, then differential evolution from concentric circles, finally locking onto a differential evolution call from a 4×4 grid with strategy rand1bin at gen 8. By gen $8$ the run is anchored at $0.6746$.
For the remaining $504$ generations greedy accepts nothing. The
displayed gen-$512$ best (\Cref{fig:case_study}) shows the consequence: $14$ of the $16$
points are reasonably spread, but the $\text{DE}$ local optimum
happens to leave one pair noticeably closer than the rest --- the
red bar in the figure --- and that single tight pair fixes
$d_{\min}$ at roughly $20\%$ of $d_{\max}$.

On CP, scipy-based accepted programs score on average $+0.35$ higher than non-scipy ones ($0.99$ vs.\ $0.64$); on MMD the sign flips, where scipy programs average $-0.14$ lower ($0.59$ vs.\ $0.73$). Both tasks have a clear
``correct'' algorithmic family: \texttt{scipy} for CP, analytic
polygon for MMD. And both have a wrong family that imposes a
hard ceiling. Whether greedy escapes depends entirely on which
family it locks into first.

\subsection{Heilbronn Triangle: the same mechanism, with stricter
feasibility}
\label{app:case_study_ht}

\paragraph{The task in one sentence.} Place $11$ points on or
inside the equilateral triangle with vertices $(0,0)$, $(1,0)$,
$(0.5,\sqrt{3}/2)$ to maximize the area of the smallest triangle
formed by any three of them.

HT differs from CP and MMD in two ways. First, the initial program
returns $11$ zeros and scores $0.0$, so every run begins with a
long stretch of invalid output --- the LLM has to write
\emph{something} that puts $11$ points inside the triangle before
any fitness signal can be received. Second, there is no analytic
attractor: every accepted improvement past the initial valid
configuration is a numerical refinement, and the score is set by a
single bottleneck triplet whose area must be enlarged without
destroying the area of any of the other $164$ triplets.

\paragraph{Healthy example --- run 6.} After $42$ generations of
invalid output, the run accepts its first valid configuration
(score $0.2445$) and then climbs through $13$ further small accepts
to $0.9510$ at generation $470$. The anchor program at gen $470$
uses \texttt{scipy.optimize.minimize} with the SLSQP method, called
from $100$ randomly perturbed initial guesses, with \emph{hard
inequality constraints} for the three triangle edges (one
constraint per point per edge) and a log-sum-exp soft-min
($\frac{1}{k}\log\sum_i e^{-k a_i}$, $k{=}1000$) over the $165$
triplet areas as the objective. The three structural choices ---
hard constraints, multi-start, smooth soft-min --- are what
distinguish this program from every stuck-HT program.

\paragraph{Stuck example --- run 5.} At gen 46 the run anchors on a
\texttt{differential\_evolution} program with bounds $[0,1]^{22}$
(the unit square, which strictly contains the triangle) and a soft
$+1000$ penalty per outside-triangle violator; anchor score is
$0.2641$. The run then freezes for $466$ further generations, during
which $465$ children evaluate to fitness $0.0$ --- DE returns at
least one out-of-triangle point and the evaluator raises a
\texttt{ValueError}. The soft penalty is the structural culprit: a
flat constant above a sharp feasibility boundary, with no gradient
telling DE which way to pull a violator back in. The single
positive-fit child ($0.2577$) is still below the anchor. The
displayed gen-$512$ best (\Cref{fig:case_study}) shows the
pathology: three of the eleven points lie on a near-straight
diagonal, and the small triangle they form is the bottleneck.

\paragraph{HT in summary.} The mechanism is the same as MMD ---
greedy commits to one algorithmic template early, and the template
itself imposes the ceiling --- but with one extra failure mode
specific to HT: when the chosen template uses soft penalties on a
hard feasibility constraint, the vast majority of mutations fail
the constraint outright and are silently rejected at score $0.0$.

\paragraph{Implication for depth-breadth.}
Across all three tasks, outcomes are gated by which algorithmic 
family the run anchors on in the initial generations. Breadth and 
depth play complementary roles here: breadth gives more parallel 
attempts at finding a viable family, while depth refines within 
the family once anchored. This explains why tasks with multiple 
near-viable families benefit from balanced allocation, whereas 
tasks with a single dominant family tolerate either extreme.

\begin{figure}[t]
    \centering
    \includegraphics[width=\linewidth]{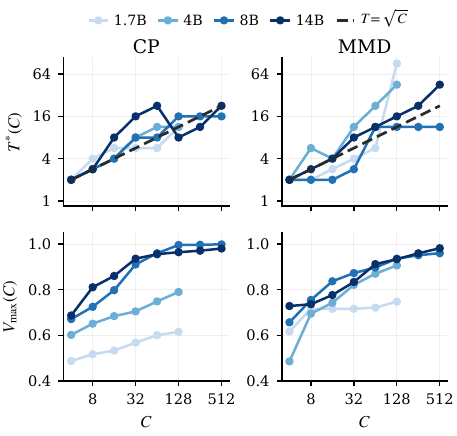}
    \caption{Cross-model allocation and performance scaling.  Top:
    near-optimal depth $T^*(C)$ as a function of total LLM call budget.
    Bottom: best attainable mean fitness $V_{\max}(C)$ at the same budgets.
    The depth curves do not form a single transferable law, while
    $V_{\max}(C)$ is substantially smoother and more capability-ordered.  
    }
    \label{fig:xmodel_scaling_tv}
\end{figure}

\begin{figure*}[t]
    \centering
    \includegraphics[width=0.8\linewidth]{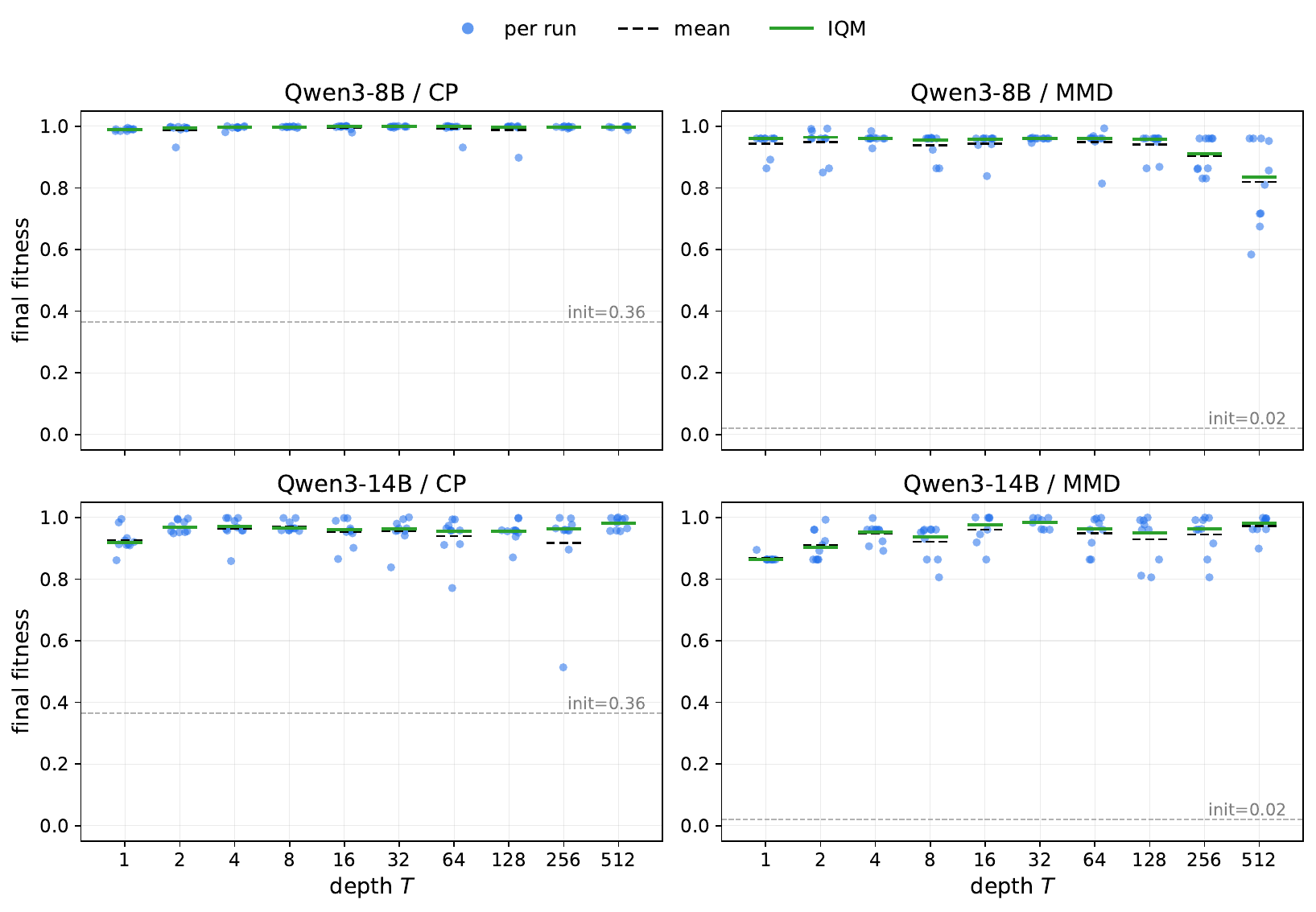}
    \caption{\textbf{Final fitness of repeated runs under an identical configuration.}
    Each dot is one run; horizontal bars mark the mean (dashed),
    and IQM (solid). For every configuration, repeated runs spread over
    a range of final fitness rather than collapsing to a single value. Grey dashed
    line: initial-program fitness.}
    \label{fig:trajectory_spread}

\end{figure*}

\begin{figure*}[t]
    \centering
    \includegraphics[width=1\linewidth]{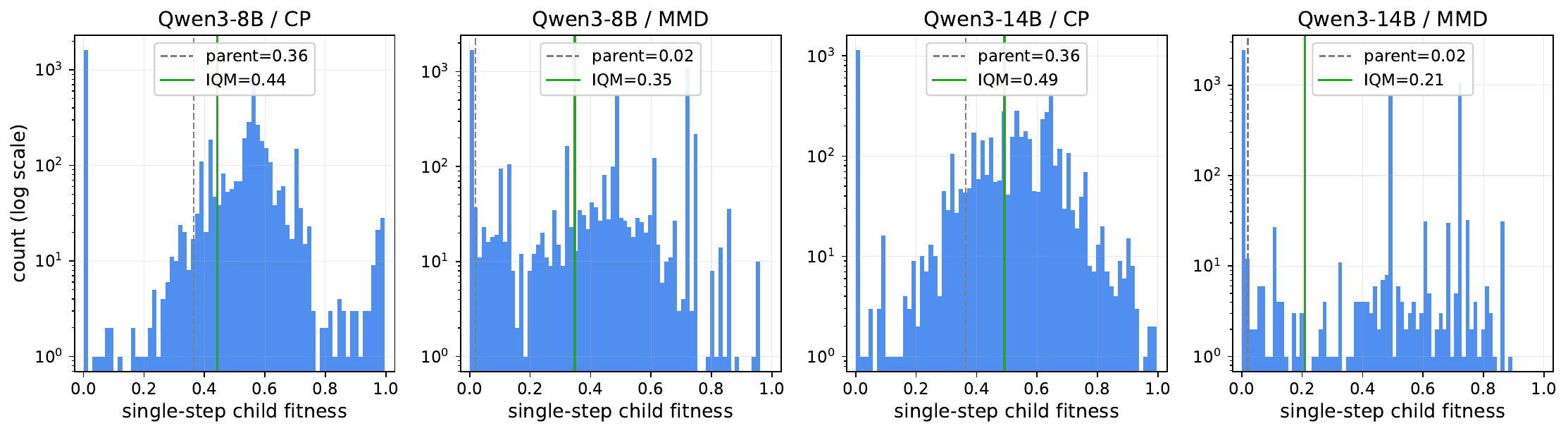}
    \caption{\textbf{Fitness of children from a single mutation step.} Distribution of
    child fitness at $T=1$, where all children share the initial program as parent
    (log count axis). One application of the LLM
    mutation operator already spreads child fitness broadly across $[0,1]$ instead of
    concentrating them near the parent; this step-level spread is the source of the
    run-to-run spread in Figure~\ref{fig:trajectory_spread}.}
    \label{fig:operator_spread}

\end{figure*}

\section{Distribution}
\label{app:distribution}

In the main text we model each self-evolution run as one sample from a latent
distribution determined by the allocation configuration
$(\text{model},\text{task},C,T)$. This appendix gives the empirical support for
that view. We run every configuration with $10$ independent runs and look
at the fitness these repeated runs produce.

Figure~\ref{fig:trajectory_spread} shows the final fitness of every run. Repeating
the same configuration does not return the same result: the different runs spread over a
range of final fitness. This is precisely the ``heterogeneous final fitness score'' described in the main text---a single run is one draw, and which draw one obtains
varies from run to run, which is why we treat a run as a sample from a latent
distribution rather than as a fixed quantity.

Figure~\ref{fig:operator_spread} traces this spread to its source. At $T=1$ all
children are generated from the same parent (the initial program), so their
fitness scores show the effect of a single mutation step. Even one step already spreads
the child fitness broadly across $[0,1]$ instead of concentrating them near the
parent. The run-to-run spread in Figure~\ref{fig:trajectory_spread} is the
accumulation of this step-level spread over $T$ generations.

Because each configuration yields a spread rather than a single value, we
summarise it by the interquartile mean (IQM), the mean of the
central half of the runs. As Figure~\ref{fig:trajectory_spread} shows, the mean,
median, and IQM of each configuration are close, so the reported results are not
sensitive to this choice; we use IQM for robustness to occasional unusually high
or low trajectories.

\section{Depth--Breadth Regularity}
\label{app:production_law}
\subsection{Robustness to Richer Specifications}
\label{app:higher_order_fits}

\Cref{sec:production_law} claims that adding higher-order terms
to the bilinear form (\Cref{eq:bi-law}) does not materially improve
the fit. \Cref{tab:higher_order_fits} reports the $R^{2}$ of five
nested specifications on the same sub-ceiling cells ($V<0.97$) as
\Cref{tab:production_law_app}, plus an F-test of each richer
specification against the bilinear (M1) baseline.

\begin{table*}[ht]
\centering
\small
\setlength{\tabcolsep}{4pt}
\caption{Nested model comparison on the five main-experiment cells.
M0: budget-only ($c{=}0$, $a{=}b$). M1: bilinear (\Cref{eq:bi-law}).
M2/M3/M4: M1 plus $\log^{2}T$, plus $\log^{2}N$, or plus both.
$p_{\!F}$ columns give the F-test $p$-value for the richer model vs.\
M1.}
\label{tab:higher_order_fits}
\begin{tabular}{llrrrrrrrr}
\toprule
& & \multicolumn{5}{c}{$R^{2}$} & \multicolumn{3}{c}{F-test vs. M1} \\
\cmidrule(lr){3-7}\cmidrule(lr){8-10}
Task & Model & M0 & M1 & M2 & M3 & M4 & $p_{\!F}(M2)$ & $p_{\!F}(M3)$ & $p_{\!F}(M4)$ \\
\midrule
CP  & 8B  & 0.780 & 0.811 & 0.816 & 0.812 & 0.819 & 0.44 & 0.86 & 0.60 \\
CP  & 14B & 0.765 & 0.785 & 0.830 & 0.806 & 0.833 & \textbf{0.002} & \textbf{0.04} & \textbf{0.005} \\
MMD & 8B  & 0.744 & 0.916 & 0.917 & 0.917 & 0.917 & 0.62 & 0.50 & 0.78 \\
MMD & 14B & 0.754 & 0.869 & 0.876 & 0.870 & 0.877 & 0.13 & 0.81 & 0.29 \\
HT  & 8B  & $0.526$ & $0.853$ & $0.900$ & $0.856$ & $0.903$ & \textbf{0.000} & $0.36$ & \textbf{0.000} \\
\bottomrule
\end{tabular}
\end{table*}

The bilinear form is preferred on three of the five cells
(CP 8B, MMD 8B, MMD 14B): F-tests against M2, M3, and M4 are all
$p>0.1$ and the $R^{2}$ gain from any richer specification is at
most $+0.008$. CP 14B and HT 8B are the exceptions --- on both,
adding $\log^{2}T$ is significant ($p\le 0.002$) and yields a
$\Delta R^{2}$ of $+0.045$ and $+0.047$ respectively. On those two
rows a richer specification fits noticeably better, though the
fitted $c$ remains small in magnitude in both bilinear and richer
forms.

The qualitative regime classification through $c$, however, is robust
to spec choice on all five cells. \Cref{tab:higher_order_c_stability}
reports the interaction coefficient under each spec.

\begin{table*}[ht]
\centering
\small
\caption{Interaction coefficient $c$ under each specification.
Magnitudes shift but the regime (corner-favoring vs. interior-favoring)
does not: CP and HT rows stay near zero, MMD rows stay clearly negative.}
\label{tab:higher_order_c_stability}
\begin{tabular}{llrrrr}
\toprule
Task & Model & $c$ (M1) & $c$ (M2) & $c$ (M3) & $c$ (M4) \\
\midrule
CP  & 8B  & $-0.027$ & $-0.062$ & $-0.033$ & $-0.110$ \\
CP  & 14B & $+0.007$ & $+0.061$ & $-0.045$ & $+0.031$ \\
MMD & 8B  & $-0.106$ & $-0.100$ & $-0.115$ & $-0.111$ \\
MMD & 14B & $-0.057$ & $-0.078$ & $-0.053$ & $-0.091$ \\
HT  & 8B  & $+0.012$ & $-0.040$ & $+0.024$ & $-0.063$ \\
\bottomrule
\end{tabular}
\end{table*}

On the MMD rows $c$ is essentially unchanged across specs
(within $\pm 0.02$ of the bilinear estimate), so the interior-optimum
classification is unambiguous. On the CP and HT rows $c$ varies more
in magnitude and even flips sign across specs, but every estimate
has $|c|<0.12$ and these rows are already not statistically
distinguishable from zero under M1 (\Cref{tab:production_law_app});
the flip reflects noise around the $c\approx 0$ limit rather than a
genuine regime change, consistent with the discussion in
\Cref{app:boundary}. We therefore retain the bilinear form for the
narrative in \Cref{sec:production_law}: it is the simplest
specification that captures the regime structure, and the structure
itself does not change under the richer specifications.

\subsection{Plateau Width and Online Search}
\label{app:plateau_width}

The derivation below is an algebraic consequence of the
bilinear ansatz of \Cref{eq:bi-law}, not an independently fitted
scaling law. The closed-form $T^{*}(C)\propto\sqrt{C}$ that appears
en route to the plateau-width result inherits its $\tfrac{1}{2}$
exponent from the symmetry of the $\log T\log N$ cross-term; the
empirical $T^{*}(C)$ curves in \Cref{fig:xmodel_scaling_tv} are
noisy and non-monotone across models and do not pin this exponent
down independently. We therefore use the closed form only as a
stepping stone to the plateau-width result, which is the substantive
claim.

Near the optimum $T^*(C)$, we expand $\log(1-V)$ as a quadratic in
$\log T$ along the budget slice $C = TN$. Substituting
$\log N = \log C - \log T$ into \Cref{eq:bi-law} gives
\begin{equation}
\begin{split}
    \log(1{-}V) &= (\beta_0 + b\log C) + (a {-} b {+} c\log C)\log T \\
              &\quad - c(\log T)^2,
\end{split}
\end{equation}
which is a quadratic in $\log T$ with curvature $-2c$ and vertex at
\begin{equation}
    \log T^*(C) = \frac{a-b+c\log C}{2c} = \frac{a-b}{2c} +
    \frac{\log C}{2}.
\end{equation}
The fitness gap at distance $\delta = \log T - \log T^*$ from the
optimum satisfies
\begin{equation}
    \log(1-V) - \log(1-V^*) = -c\,\delta^2,
\end{equation}
so the plateau half-width within $\Delta$ of the optimal log fitness
gap is
\begin{equation}\label{eq:plateau_width}
    |\delta| \leq \sqrt{\frac{\Delta}{|c|}}.
\end{equation}
When $|c|$ is small the plateau is wide and missing $T^*$ costs
little fitness; when $|c|$ is large the ridge is sharp but the
stronger curvature signal makes $T^*$ easier to identify from online
feedback. In either case a cheap online searcher achieves near-maximum in-sweep fitness, justifying the approach of
\Cref{sec:MAB_for_trajectory_allocation}.
\Cref{fig:xmodel_scaling_tv} shows this empirically: the
estimated $T^{*}(C)$ curves are noisy and non-monotone across the
four Qwen3 sizes, yet $V_{\max}(C)$ on the same budget axis is
smooth and capability-ordered. The fitness \emph{value} at the
optimum is therefore stable even when its precise \emph{location}
is not --- the surface is searchable in the sense that what online
search achieves does not depend sensitively on identifying $T^{*}$
exactly.

\subsection{Boundary of the Regime}\label{app:boundary}

The CP row sign-flips between 8B ($c=-0.027$, $p=0.49$) and
14B ($c=+0.007$, $p=0.77$). Both estimates are statistically consistent with zero, so the flip is noise around the
$c\approx 0$ limit rather than a regime change; it is the
continuous-$|c|$ framing of \Cref{sec:production_law} that is
informative on CP, not the sign of $c$ itself.

The lower rows of \Cref{tab:production_law} mark where the regularity stops
applying. Well-fit small-model rows (CP~4B $R^2{=}0.87$, CP~1.7B
$R^2{=}0.72$) have $|c|\le0.024$, exhibiting the same geometry in
attenuated form. Poorly-fit rows (CP~Llama $R^2{=}0.33$, MMD~1.7B
$R^2{=}0.36$) never enter the sub-ceiling regime: weak models do not
produce mutations structured enough for the depth--breadth tradeoff
to become active, the same capability bottleneck identified in
\Cref{sec:envelope}.

\begin{figure}[ht]
\centering
\includegraphics[width=\linewidth]{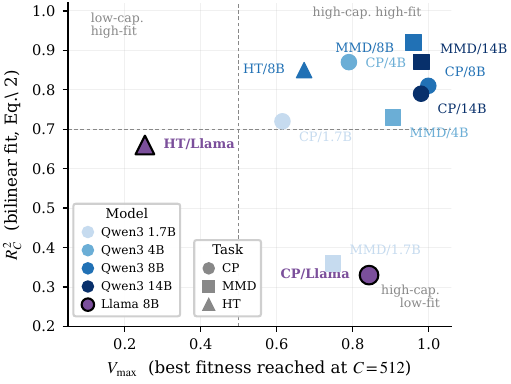}
\caption{Bilinear-fit quality $R^{2}_{C}$ versus best fitness
$V_{\max}$ (at $C{=}512$) across the eleven (model, task) cells.
Qwen3 cells (blue ramp) cluster in the upper-right; Llama-3.1-8B
cells (purple, black outline) sit below the $R^{2}=0.7$ threshold
even when $V_{\max}$ is high (CP/Llama at $V_{\max}=0.843$ but
$R^{2}_{C}=0.33$). The bilinear regularity's lower boundary is therefore
not strictly a capability floor.}
\label{fig:model_family_split}
\end{figure}

One nuance: the form-mismatch cells are not strictly
capability failures. \Cref{fig:model_family_split} plots
$R^{2}_{C}$ against $V_{\max}$ across the eleven (model, task) cells
we fit; the cells where $R^{2}_{C}<0.7$ are
\textbf{CP/Llama, HT/Llama}, and \textbf{MMD/1.7B}. MMD/1.7B is a
classic capability-floor failure ($V_{\max}=0.748$, only marginally
above best-of-$N$), but CP/Llama sits at $V_{\max}=0.843$ which is a
non-trivial absolute level and still fails the bilinear fit
($R^{2}_{C}=0.33$). The shared signature is model family, not raw
fitness: Llama's proposals lack the depth--breadth interaction
structure the bilinear form detects even when Llama itself reaches
competitive fitness. The bilinear regularity is therefore best read as a
property of the Qwen3 family's mutation distribution, with the lower
boundary of its domain being more model-family-specific than purely
capability-driven. The Spearman correlations between $V_{\max}$,
$R^{2}_{C}$, and $-\log_{10}p$ across all cells are individually
weak ($\rho\le 0.45$), consistent with this: the three statistics do
not collapse to one capability axis.

\begin{table*}[t]
\centering
\small
\caption{Bilinear fitness-gap model (\Cref{eq:bi-law}) fit on
sub-ceiling cells of the $C=512$ sweeps. Permutation $p$-values
test $c \neq 0$ ($10{,}000$ shuffles). $^{***}p<0.001$,
$^{**}p<0.01$, $^{*}p<0.05$. \textbf{Top block (main-experiment
models):} bilinear form fits cleanly across all three tasks
($R^2 \in [0.75, 0.92]$), with task-dependent $c$ --- CP has no
significant interaction (broad plateau), while MMD and HT have
significant negative interactions (interior ridge).
\textbf{Bottom block (smaller-model regime):} well-fit rows
($R^2 \ge 0.72$) all satisfy $|c| \le 0.024$, consistent with a
flattened version of the same surface; weak-model configurations
(MMD~1.7B, CP~Llama) fail to enter the sub-ceiling
regime and the form does not describe them ($R^2 \le 0.36$);
significance stars for these rows should be disregarded.}
\label{tab:production_law_app}
\begin{tabular}{llrrrrrr}
\toprule
Task & Model & $\beta_0$ & $a$ & $b$ & $c$ & $p(c)$ & $R^2$ \\
\midrule
CP  & 8B    & -0.020 & -0.602 & -0.496 & $-0.027$           & 0.49     & 0.81 \\
CP  & 14B   & -0.561 & -0.442 & -0.373 & $+0.007$           & 0.77     & 0.79 \\
MMD & 8B    & -0.590 & -0.208 & -0.290 & $-0.106^{***}$     & $1.2\!\times\!10^{-11}$ & 0.92 \\
MMD & 14B   & -0.641 & -0.342 & -0.238 & $-0.057^{***}$     & $5.0\!\times\!10^{-4}$  & 0.87 \\
HT  & 8B    & +0.592 & -0.363 & -0.168 & $-0.012^{**}$      & 0.004    & 0.85 \\
\midrule
CP  & 1.7B  & -0.498 & -0.076 & -0.088 & $+0.001$           & 0.78     & 0.72 \\
CP  & 4B    & -0.524 & -0.190 & -0.157 & $-0.017^{*}$       & 0.045    & 0.87 \\
CP  & Llama & -0.917 & -0.017 & -0.101 & $-0.010$           & 0.53     & 0.33$^\dagger$ \\
MMD & 1.7B  & -1.131 & -0.035 & -0.030 & $+0.000$           & 0.98     & 0.36$^\dagger$ \\
MMD & 4B    & -0.223 & -0.418 & -0.248 & $+0.024$           & 0.30     & 0.73 \\
HT  & Llama & +0.013 & -0.026 & -0.006 & $-0.009$           & $5.0\!\times\!10^{-4}$  & 0.66$^\dagger$ \\
\bottomrule
\end{tabular}
\end{table*}

\begin{table*}[t]
\centering
\caption{Pairwise best-fitness comparisons: each block adapts one parent sampling protocol (\emph{greedy protocol} with different breadth $N$ or \emph{island protocols} including OpenEvolve, CodeEvolve, and ShinkaEvolve) into a baseline self-evolving process as described in \Cref{alg:self_evolving} and \texttt{BaSE} as described in \Cref{alg:base}, where \texttt{BaSE} includes three bandit algorithms (UCB, EXP3.P, and Thompson sampling). The best mean fitness score in each block is \textbf{bolded}.}
\label{tab:pairwise_fitness}
\setlength{\tabcolsep}{2.5pt}
\begin{adjustbox}{width=\textwidth,center}
\begin{tabular}{ll l cccc ccc}
\toprule
\textbf{Model} & \textbf{Task} & \textbf{Method}
& \textbf{Greedy-1}
& \textbf{Greedy-2}
& \textbf{Greedy-4}
& \textbf{Greedy-8}
& \textbf{OpenEvolve}
& \textbf{CodeEvolve}
& \textbf{ShinkaEvolve} \\
\midrule

\multirow{12}{*}{Qwen3-8B}
& \multirow{4}{*}{CP}
& Baseline
& 0.9969\,{\scriptsize$\pm$\,.0011}
& 0.9974\,{\scriptsize$\pm$\,.0007}
& 0.9884\,{\scriptsize$\pm$\,.0093}
& 0.9912\,{\scriptsize$\pm$\,.0066}
& 0.8279\,{\scriptsize$\pm$\,.0145}
& 0.7692\,{\scriptsize$\pm$\,.0202}
& \textbf{0.9986\,{\scriptsize$\pm$\,.0003}} \\
& & UCB
& \textbf{0.9985\,{\scriptsize$\pm$\,.0011}}
& \textbf{0.9985\,{\scriptsize$\pm$\,.0004}}
& 0.9980\,{\scriptsize$\pm$\,.0013}
& 0.9965\,{\scriptsize$\pm$\,.0102}
& 0.8031\,{\scriptsize$\pm$\,.0509}
& 0.8031\,{\scriptsize$\pm$\,.0325}
& 0.9978\,{\scriptsize$\pm$\,.0006} \\
& & EXP3.P
& 0.9983\,{\scriptsize$\pm$\,.0013}
& 0.9983\,{\scriptsize$\pm$\,.0007}
& \textbf{0.9987\,{\scriptsize$\pm$\,.0027}}
& 0.9993\,{\scriptsize$\pm$\,.0059}
& 0.7946\,{\scriptsize$\pm$\,.0370}
& \textbf{0.8191\,{\scriptsize$\pm$\,.0280}}
& 0.9983\,{\scriptsize$\pm$\,.0007} \\
& & Thompson
& 0.9969\,{\scriptsize$\pm$\,.0100}
& \textbf{0.9985\,{\scriptsize$\pm$\,.0004}}
& \textbf{0.9987\,{\scriptsize$\pm$\,.0096}}
& \textbf{1.0003\,{\scriptsize$\pm$\,.0174}}
& \textbf{0.8337\,{\scriptsize$\pm$\,.0223}}
& \textbf{0.8191\,{\scriptsize$\pm$\,.0291}}
& 0.9985\,{\scriptsize$\pm$\,.0004} \\
\cmidrule(lr){2-10}

& \multirow{4}{*}{MMD}
& Baseline
& 0.8197\,{\scriptsize$\pm$\,.0427}
& 0.9047\,{\scriptsize$\pm$\,.0177}
& \textbf{0.9407\,{\scriptsize$\pm$\,.0121}}
& 0.9477\,{\scriptsize$\pm$\,.0153}
& 0.9561\,{\scriptsize$\pm$\,.0143}
& 0.8414\,{\scriptsize$\pm$\,.0689}
& 0.9099\,{\scriptsize$\pm$\,.0240} \\
& & UCB
& \textbf{0.9603\,{\scriptsize$\pm$\,.0129}}
& \textbf{0.9603\,{\scriptsize$\pm$\,.0114}}
& 0.8930\,{\scriptsize$\pm$\,.0156}
& 0.9603\,{\scriptsize$\pm$\,.0053}
& 0.9307\,{\scriptsize$\pm$\,.0380}
& \textbf{0.9398\,{\scriptsize$\pm$\,.0839}}
& \textbf{0.9630\,{\scriptsize$\pm$\,.0471}} \\
& & EXP3.P
& 0.9595\,{\scriptsize$\pm$\,.0386}
& \textbf{0.9603\,{\scriptsize$\pm$\,.0217}}
& 0.9119\,{\scriptsize$\pm$\,.0440}
& \textbf{0.9658\,{\scriptsize$\pm$\,.0102}}
& 0.9442\,{\scriptsize$\pm$\,.0446}
& 0.9090\,{\scriptsize$\pm$\,.0860}
& 0.9602\,{\scriptsize$\pm$\,.0363} \\
& & Thompson
& \textbf{0.9603\,{\scriptsize$\pm$\,.0546}}
& 0.9215\,{\scriptsize$\pm$\,.0460}
& 0.8634\,{\scriptsize$\pm$\,.0440}
& 0.9603\,{\scriptsize$\pm$\,.0035}
& \textbf{0.9605\,{\scriptsize$\pm$\,.0347}}
& 0.8653\,{\scriptsize$\pm$\,.0833}
& 0.9341\,{\scriptsize$\pm$\,.0256} \\
\cmidrule(lr){2-10}

& \multirow{4}{*}{HT}
& Baseline
& 0.6441\,{\scriptsize$\pm$\,.0723}
& 0.6780\,{\scriptsize$\pm$\,.0718}
& \textbf{0.5502\,{\scriptsize$\pm$\,.0666}}
& 0.5488\,{\scriptsize$\pm$\,.0814}
& 0.6061\,{\scriptsize$\pm$\,.0768}
& 0.5168\,{\scriptsize$\pm$\,.0627}
& 0.7379\,{\scriptsize$\pm$\,.0373} \\
& & UCB
& 0.7250\,{\scriptsize$\pm$\,.0684}
& 0.8530\,{\scriptsize$\pm$\,.0963}
& 0.5238\,{\scriptsize$\pm$\,.0592}
& 0.6498\,{\scriptsize$\pm$\,.0573}
& \textbf{0.8624\,{\scriptsize$\pm$\,.1464}}
& 0.6912\,{\scriptsize$\pm$\,.0364}
& \textbf{0.7528\,{\scriptsize$\pm$\,.1188}} \\
& & EXP3.P
& 0.7421\,{\scriptsize$\pm$\,.1000}
& 0.8653\,{\scriptsize$\pm$\,.1068}
& 0.4633\,{\scriptsize$\pm$\,.1078}
& 0.6780\,{\scriptsize$\pm$\,.1057}
& 0.7598\,{\scriptsize$\pm$\,.1518}
& \textbf{0.7164\,{\scriptsize$\pm$\,.0483}}
& 0.6642\,{\scriptsize$\pm$\,.1000} \\
& & Thompson
& \textbf{0.8104\,{\scriptsize$\pm$\,.1785}}
& \textbf{0.8736\,{\scriptsize$\pm$\,.1737}}
& 0.3276\,{\scriptsize$\pm$\,.1426}
& \textbf{0.7898\,{\scriptsize$\pm$\,.1225}}
& 0.6503\,{\scriptsize$\pm$\,.1571}
& \textbf{0.7164\,{\scriptsize$\pm$\,.0593}}
& 0.7132\,{\scriptsize$\pm$\,.1013} \\
\midrule

\multirow{12}{*}{Llama}
& \multirow{4}{*}{CP}
& Baseline
& 0.7034\,{\scriptsize$\pm$\,.0657}
& 0.6325\,{\scriptsize$\pm$\,.0610}
& 0.6699\,{\scriptsize$\pm$\,.0591}
& 0.6584\,{\scriptsize$\pm$\,.0583}
& 0.7033\,{\scriptsize$\pm$\,.0657}
& 0.4568\,{\scriptsize$\pm$\,.0330}
& 0.8305\,{\scriptsize$\pm$\,.0362} \\
& & UCB
& 0.9451\,{\scriptsize$\pm$\,.0633}
& \textbf{0.9076\,{\scriptsize$\pm$\,.1536}}
& \textbf{0.9449\,{\scriptsize$\pm$\,.0845}}
& 0.9452\,{\scriptsize$\pm$\,.1359}
& 0.9451\,{\scriptsize$\pm$\,.0633}
& \textbf{0.7101\,{\scriptsize$\pm$\,.1457}}
& \textbf{0.9472\,{\scriptsize$\pm$\,.1325}} \\
& & EXP3.P
& 0.9484\,{\scriptsize$\pm$\,.0716}
& \textbf{0.9076\,{\scriptsize$\pm$\,.1640}}
& \textbf{0.9449\,{\scriptsize$\pm$\,.0841}}
& 0.9424\,{\scriptsize$\pm$\,.1517}
& 0.9484\,{\scriptsize$\pm$\,.0716}
& 0.3985\,{\scriptsize$\pm$\,.1222}
& 0.8983\,{\scriptsize$\pm$\,.0918} \\
& & Thompson
& \textbf{0.9533\,{\scriptsize$\pm$\,.0706}}
& \textbf{0.9076\,{\scriptsize$\pm$\,.1743}}
& 0.9407\,{\scriptsize$\pm$\,.0989}
& \textbf{0.9511\,{\scriptsize$\pm$\,.1711}}
& \textbf{0.9533\,{\scriptsize$\pm$\,.0706}}
& 0.3985\,{\scriptsize$\pm$\,.1234}
& 0.9176\,{\scriptsize$\pm$\,.1140} \\
\cmidrule(lr){2-10}

& \multirow{4}{*}{MMD}
& Baseline
& 0.6388\,{\scriptsize$\pm$\,.1030}
& 0.5182\,{\scriptsize$\pm$\,.1011}
& 0.6576\,{\scriptsize$\pm$\,.0878}
& 0.5863\,{\scriptsize$\pm$\,.1170}
& 0.6378\,{\scriptsize$\pm$\,.1024}
& 0.2315\,{\scriptsize$\pm$\,.0847}
& 0.7253\,{\scriptsize$\pm$\,.0453} \\
& & UCB
& \textbf{0.7978\,{\scriptsize$\pm$\,.0575}}
& \textbf{0.8777\,{\scriptsize$\pm$\,.0856}}
& 0.7386\,{\scriptsize$\pm$\,.0235}
& \textbf{0.8643\,{\scriptsize$\pm$\,.1214}}
& \textbf{0.7978\,{\scriptsize$\pm$\,.0575}}
& \textbf{0.4906\,{\scriptsize$\pm$\,.2244}}
& \textbf{0.8056\,{\scriptsize$\pm$\,.0639}} \\
& & EXP3.P
& \textbf{0.7978\,{\scriptsize$\pm$\,.0542}}
& 0.8652\,{\scriptsize$\pm$\,.1024}
& 0.7882\,{\scriptsize$\pm$\,.0579}
& \textbf{0.8643\,{\scriptsize$\pm$\,.1659}}
& \textbf{0.7978\,{\scriptsize$\pm$\,.0542}}
& \textbf{0.4906\,{\scriptsize$\pm$\,.2244}}
& 0.7589\,{\scriptsize$\pm$\,.0475} \\
& & Thompson
& \textbf{0.7978\,{\scriptsize$\pm$\,.0448}}
& 0.7115\,{\scriptsize$\pm$\,.1108}
& \textbf{0.8830\,{\scriptsize$\pm$\,.0504}}
& 0.8608\,{\scriptsize$\pm$\,.1540}
& \textbf{0.7978\,{\scriptsize$\pm$\,.0448}}
& \textbf{0.4906\,{\scriptsize$\pm$\,.2244}}
& 0.7644\,{\scriptsize$\pm$\,.0348} \\
\cmidrule(lr){2-10}

& \multirow{4}{*}{HT}
& Baseline
& \textbf{0.0192\,{\scriptsize$\pm$\,.0090}}
& \textbf{0.1722\,{\scriptsize$\pm$\,.0699}}
& 0.2538\,{\scriptsize$\pm$\,.0762}
& \textbf{0.1502\,{\scriptsize$\pm$\,.0465}}
& 0.0000\,{\scriptsize$\pm$\,.0000}
& 0.0000\,{\scriptsize$\pm$\,.0000}
& \textbf{0.1512\,{\scriptsize$\pm$\,.0437}} \\
& & UCB
& 0.0000\,{\scriptsize$\pm$\,.0000}
& 0.1562\,{\scriptsize$\pm$\,.0752}
& 0.2561\,{\scriptsize$\pm$\,.1196}
& 0.0898\,{\scriptsize$\pm$\,.0398}
& 0.0000\,{\scriptsize$\pm$\,.0000}
& 0.0000\,{\scriptsize$\pm$\,.0000}
& 0.0274\,{\scriptsize$\pm$\,.0129} \\
& & EXP3.P
& 0.0000\,{\scriptsize$\pm$\,.0000}
& 0.1293\,{\scriptsize$\pm$\,.0918}
& 0.2729\,{\scriptsize$\pm$\,.1851}
& 0.0898\,{\scriptsize$\pm$\,.0491}
& 0.0000\,{\scriptsize$\pm$\,.0000}
& 0.0000\,{\scriptsize$\pm$\,.0000}
& 0.0274\,{\scriptsize$\pm$\,.0128} \\
& & Thompson
& 0.0000\,{\scriptsize$\pm$\,.0000}
& 0.1315\,{\scriptsize$\pm$\,.0943}
& \textbf{0.4387\,{\scriptsize$\pm$\,.2134}}
& 0.0898\,{\scriptsize$\pm$\,.0412}
& 0.0000\,{\scriptsize$\pm$\,.0000}
& 0.0000\,{\scriptsize$\pm$\,.0000}
& 0.0274\,{\scriptsize$\pm$\,.0129} \\
\midrule

\multirow{8}{*}{Qwen3-14B}
& \multirow{4}{*}{CP}
& Baseline
& 0.9802\,{\scriptsize$\pm$\,.0060}
& 0.9181\,{\scriptsize$\pm$\,.0421}
& 0.9536\,{\scriptsize$\pm$\,.0106}
& 0.9394\,{\scriptsize$\pm$\,.0191}
& 0.8577\,{\scriptsize$\pm$\,.0228}
& 0.8768\,{\scriptsize$\pm$\,.0134}
& 0.9981\,{\scriptsize$\pm$\,.0003} \\
& & UCB
& 0.9972\,{\scriptsize$\pm$\,.0074}
& \textbf{0.9832\,{\scriptsize$\pm$\,.0120}}
& \textbf{0.9955\,{\scriptsize$\pm$\,.0186}}
& 0.9835\,{\scriptsize$\pm$\,.0170}
& 0.8223\,{\scriptsize$\pm$\,.0422}
& 0.9014\,{\scriptsize$\pm$\,.0655}
& 0.9974\,{\scriptsize$\pm$\,.0011} \\
& & EXP3.P
& 0.9967\,{\scriptsize$\pm$\,.0114}
& 0.9743\,{\scriptsize$\pm$\,.0121}
& 0.9722\,{\scriptsize$\pm$\,.0182}
& 0.9824\,{\scriptsize$\pm$\,.0190}
& 0.8532\,{\scriptsize$\pm$\,.0650}
& 0.8303\,{\scriptsize$\pm$\,.0677}
& \textbf{0.9983\,{\scriptsize$\pm$\,.0017}} \\
& & Thompson
& \textbf{1.0003\,{\scriptsize$\pm$\,.0143}}
& 0.9728\,{\scriptsize$\pm$\,.0086}
& 0.9576\,{\scriptsize$\pm$\,.0071}
& \textbf{0.9935\,{\scriptsize$\pm$\,.0164}}
& \textbf{0.8701\,{\scriptsize$\pm$\,.0701}}
& \textbf{0.9686\,{\scriptsize$\pm$\,.0959}}
& 0.9969\,{\scriptsize$\pm$\,.0007} \\
\cmidrule(lr){2-10}

& \multirow{4}{*}{MMD}
& Baseline
& 0.9735\,{\scriptsize$\pm$\,.0101}
& 0.9450\,{\scriptsize$\pm$\,.0194}
& 0.9288\,{\scriptsize$\pm$\,.0249}
& 0.9481\,{\scriptsize$\pm$\,.0173}
& \textbf{0.9949\,{\scriptsize$\pm$\,.0012}}
& 0.9934\,{\scriptsize$\pm$\,.0021}
& \textbf{0.9924\,{\scriptsize$\pm$\,.0037}} \\
& & UCB
& 0.9955\,{\scriptsize$\pm$\,.0366}
& \textbf{0.9915\,{\scriptsize$\pm$\,.0093}}
& 0.9603\,{\scriptsize$\pm$\,.0192}
& 0.9603\,{\scriptsize$\pm$\,.0420}
& 0.9601\,{\scriptsize$\pm$\,.0261}
& 0.9999\,{\scriptsize$\pm$\,.0266}
& 0.9478\,{\scriptsize$\pm$\,.0367} \\
& & EXP3.P
& 0.9907\,{\scriptsize$\pm$\,.0370}
& \textbf{0.9915\,{\scriptsize$\pm$\,.0141}}
& \textbf{0.9797\,{\scriptsize$\pm$\,.0168}}
& \textbf{0.9606\,{\scriptsize$\pm$\,.0343}}
& 0.9890\,{\scriptsize$\pm$\,.0148}
& \textbf{1.0000\,{\scriptsize$\pm$\,.0341}}
& 0.9888\,{\scriptsize$\pm$\,.0158} \\
& & Thompson
& \textbf{0.9966\,{\scriptsize$\pm$\,.0204}}
& \textbf{0.9915\,{\scriptsize$\pm$\,.0141}}
& 0.9660\,{\scriptsize$\pm$\,.0154}
& 0.9603\,{\scriptsize$\pm$\,.0463}
& 0.9912\,{\scriptsize$\pm$\,.0118}
& \textbf{1.0000\,{\scriptsize$\pm$\,.0363}}
& 0.9665\,{\scriptsize$\pm$\,.0167} \\

\bottomrule
\end{tabular}
\end{adjustbox}
\end{table*}

\begin{table*}[t]
    \centering
    \caption{Minimum generation (Gen.) and cumulative FLOPs ($\times 10^{15}$) required for $\geq 90\%$ of replicates to reach thresholds ($\geq \tau$) with three LLM models. Unreached thresholds are denoted as ``---''. The minimum generation and FLOPs for each threshold are \textbf{bolded}.}
    \label{tab:threshold_iteration_flops_complete}
    \begin{adjustbox}{width=\linewidth,center}
    {\scriptsize
    \begin{tabular}{@{}llr *{4}{rr} *{4}{rr}@{}}
    \toprule
    \textbf{Model} & \textbf{Task} & $\boldsymbol{\tau}$ & \multicolumn{8}{c}{\textbf{Baselines}} & \multicolumn{8}{c}{\textbf{\texttt{BaSE}}} \\
    \cmidrule(lr){4-11}\cmidrule(lr){12-19}
    & & & \multicolumn{2}{c}{Greedy} & \multicolumn{2}{c}{OpenEvolve} & \multicolumn{2}{c}{CodeEvolve} & \multicolumn{2}{c}{ShinkaEvolve} & \multicolumn{2}{c}{Random} & \multicolumn{2}{c}{UCB} & \multicolumn{2}{c}{EXP3} & \multicolumn{2}{c}{Thompson} \\
    \cmidrule(lr){4-5}\cmidrule(lr){6-7}\cmidrule(lr){8-9}\cmidrule(lr){10-11}
    \cmidrule(lr){12-13}\cmidrule(lr){14-15}\cmidrule(lr){16-17}\cmidrule(lr){18-19}
    & &  & Gen. & FLOPs & Gen. & FLOPs & Gen. & FLOPs & Gen. & FLOPs 
      & Gen. & FLOPs & Gen. & FLOPs & Gen. & FLOPs & Gen. & FLOPs \\
    \midrule

    \multirow{7}{*}{Qwen3-14B} & \multirow{4}{*}{CP} 
    & 0.90 & 46 & 50.19 & --- & --- & --- & --- & \textbf{8} & \textbf{9.41} & 68 & 80.33 & 60 & 72.66 & 58 & 67.44 & 28 & 30.28 \\
    & & 0.95 & 186 & 134.13 & --- & --- & --- & --- & \textbf{39} & \textbf{43.71} & 120 & 142.52 & 60 & 72.66 & 122 & 139.76 & 49 & 51.43 \\
    & & 0.99 & --- & --- & --- & --- & --- & --- & 146 & 114.73 & 137 & 163.87 & \textbf{60} & \textbf{72.66} & 137 & 155.09 & 295 & 284.78 \\
    & & 0.999 & --- & --- & --- & --- & --- & --- & --- & --- & --- & --- & --- & --- & --- & --- & \textbf{295} & \textbf{284.78} \\
    \cmidrule(lr){2-19}
    & \multirow{3}{*}{MMD} & 0.80 & 257 & 289.20 & 80 & 92.46 & --- & --- & 34 & 40.02 & 59 & 54.39 & \textbf{11} & \textbf{9.95} & 75 & 70.48 & \textbf{11} & \textbf{9.95} \\
    & & 0.90 & 372 & 406.99 & \textbf{146} & 168.19 & --- & --- & 165 & 189.18 & 175 & 171.16 & \textbf{146} & \textbf{148.08} & 226 & 229.18 & 166 & 160.66 \\
    & & 0.95 & 418 & 451.65 & 259 & 291.33 & --- & --- & 210 & 239.02 & 226 & 224.90 & \textbf{146} & \textbf{148.08} & 328 & 331.98 & 178 & 171.35 \\
    \midrule

    \multirow{7}{*}{Llama} & \multirow{2}{*}{CP} 
    & 0.60 & 262 & 55.30 & --- & --- & --- & --- & 30 & 5.67 & 18 & 2.55 & \textbf{5} & \textbf{0.87} & 12 & 1.78 & \textbf{5} & \textbf{0.87} \\
    & & 0.80 & 262 & 55.30 & --- & --- & --- & --- & --- & --- & 18 & 2.55 & \textbf{5} & \textbf{0.87} & 12 & 1.78 & \textbf{5} & \textbf{0.87} \\
    \cmidrule(lr){2-19}
    & \multirow{4}{*}{MMD} & 0.60 & \textbf{215} & 58.25 & --- & --- & --- & --- & --- & --- & 417 & 34.75 & 321 & \textbf{18.99} & 321 & 27.61 & 257 & 20.55 \\
    & & 0.70 & \textbf{215} & 58.25 & --- & --- & --- & --- & --- & --- & 451 & 39.13 & 324 & \textbf{19.44} & 324 & 29.20 & 260 & 20.88 \\
    & & 0.80 & --- & --- & --- & --- & --- & --- & --- & --- & --- & --- & 371 & \textbf{22.74} & 435 & 52.97 & \textbf{323} & 33.49 \\
    & & 0.90 & --- & --- & --- & --- & --- & --- & --- & --- & --- & --- & \textbf{400} & \textbf{38.13} & --- & --- & --- & --- \\
    \cmidrule(lr){2-19}
    & \multirow{1}{*}{HT} 
    & 0.60 & --- & --- & --- & --- & --- & --- & --- & --- & --- & --- & --- & --- & --- & --- & --- & --- \\
    \bottomrule
    \end{tabular}
    }
    \end{adjustbox}
\end{table*}

\section{Deferred \texttt{BaSE} Experiments}

\subsection{Pairwise Fitness Comparisons}
\label{sec:pairwise_fitness}

\Cref{tab:pairwise_fitness} reports pairwise comparisons under the same LLM-call budget $C$. For each parent sampling protocol, we compare its original self-evolving version with a \texttt{BaSE} variant that uses the same prompt generator, i.e., the same parent sampling protocol with identical configuration, so as to focus on the effect of adaptive trajectory allocation. 

Overall, \texttt{BaSE} improves the best fitness score in most settings, especially when the original protocol has substantial remaining headroom or high variance across trajectories. Under the greedy $T{=}512$ protocol, \texttt{BaSE} improves Qwen3-8B MMD from $0.8197$ to $0.9603$ and HT from $0.6441$ to $0.8104$, while also improving Llama CP from $0.7034$ to $0.9533$ and MMD from $0.6388$ to $0.7978$. Similar gains appear under greedy $T{=}256$, where Qwen3-8B HT increases from $0.6780$ to $0.8736$, and Llama MMD increases from $0.5182$ to $0.8777$.

The improvement is smaller when the original protocol already saturates the task. For example, on CP, several methods already achieve near-ceiling fitness scores, leaving little room for further gains. This is visible for Qwen3-8B ShinkaEvolve on CP, where the original protocol reaches $0.9986$ and \texttt{BaSE} variants remain at a similar level. In contrast, \texttt{BaSE} yields clearer gains in harder or more unstable settings by allocating more computation to trajectories that appear more promising. For instance, when using the CodeEvolve prompt generator, \texttt{BaSE} improves Qwen3-8B HT from $0.5168$ to $0.7164$, Llama MMD from $0.2315$ to $0.4906$, and Qwen3-14B CP from $0.8768$ to $0.9686$.

In some cases, especially when the original breadth-depth allocation is already strong, or when all candidate trajectories have limited potential, \texttt{BaSE} can underperform the original protocol. For example, Qwen3-8B greedy with $T{=}128,N{=}4$ performs better than \texttt{BaSE} on MMD and HT, and Llama HT remains difficult across several parent sampling protocols. These results suggest that \texttt{BaSE} is most useful when the candidate runs contain heterogeneous trajectory quality that can be exploited by adaptive allocation.

\subsection{Full Threshold Comparison}
\label{sec:full_threshold_comparison}

\Cref{tab:threshold_iteration_flops_complete} reports the complete threshold-reaching results across three backbone models. Overall, the same pattern observed in the main table remains consistent: \texttt{BaSE} variants, especially UCB and Thompson sampling, reach target fitness levels with fewer generations and lower cumulative FLOPs in most settings where the thresholds are attainable. On Qwen3-8B, \texttt{BaSE} dominates MMD and HT, with UCB/Thompson reaching MMD thresholds up to $\tau=0.95$ substantially earlier than greedy, while several evolution baselines fail to reach these targets. On CP, ShinkaEvolve is competitive for intermediate thresholds, but Thompson is the only method that reaches the stringent $\tau=0.999$ threshold. Similar trends hold for Qwen3-14B: \texttt{BaSE}-UCB is strongest on MMD and reaches CP $\tau=0.99$ earlier than all baselines, while Thompson is again the only method reaching CP $\tau=0.999$. For Llama, absolute performance is lower and many thresholds remain unreached, especially on HT; nevertheless, \texttt{BaSE} still provides the most reliable threshold-reaching behavior, achieving the best or lowest-FLOP results on most reachable CP and MMD thresholds. These results suggest that adaptive trajectory allocation improves sample efficiency across model scales, while its gains are most pronounced on tasks where evolutionary trajectories exhibit substantial variance and early allocation decisions matter.

\subsection{Ablation on Bandit Arm-Pool Size}
\label{sec:bandit_ablation_arms}

To study how the number of parallel trajectories affects online budget allocation, we ablate the arm-pool size $K \in \{2,5,10,20,50\}$ for \texttt{BaSE} under the same fixed generation budget of $C=512$ with configuration $T = 512, N = 1$. As shown in \Cref{tab:bandit_ablation_arms}, performance is generally strongest with a moderate number of arms rather than the largest pool. This reflects a natural cross-trajectory level breadth--depth trade-off: increasing $K$ provides more trajectory diversity, but also reduces the number of refinement steps that can be allocated to each trajectory within the fixed budget. For Qwen3-8B, MMD reaches its best score with $K=5$--$20$, while HT performs best with small or moderate arm pools, indicating that excessive breadth can leave promising trajectories under-refined. Similarly, Qwen3-14B achieves strong CP performance across several $K$ values under Thompson sampling, while MMD peaks at $K=5$. For Llama, moderate arm pools again perform best on CP and MMD, whereas all methods fail on HT, suggesting that allocation alone cannot compensate when the underlying model rarely produces viable improvements.

\begin{table*}[t]
    \centering
    \caption{\texttt{BaSE} running-max fitness at $T=512$ versus bandit arm-pool size $K \in \{2, 5, 10, 20, 50\}$. The largest mean within each (Model, Task) block is \textbf{bolded}.}
    \label{tab:bandit_ablation_arms}
    \begin{adjustbox}{width=\linewidth}
    \begin{tabular}{lllccccc}
    \toprule
    \textbf{Model} & \textbf{Task} & \textbf{Algorithm} & \textbf{$K=2$} & \textbf{$K=5$} & \textbf{$K=10$} & \textbf{$K=20$} & \textbf{$K=50$} \\
    \midrule
    \multirow{9}{*}{Qwen3-8B} & \multirow{3}{*}{CP} & UCB & 0.9981\,{\scriptsize$\pm$\,.0004} & 0.9985\,{\scriptsize$\pm$\,.0024} & \textbf{0.9985\,{\scriptsize$\pm$\,.0011}} & 0.9981\,{\scriptsize$\pm$\,.0005} & 0.9937\,{\scriptsize$\pm$\,.0001} \\
     &  & EXP3.P & 0.9981\,{\scriptsize$\pm$\,.0003} & 0.9983\,{\scriptsize$\pm$\,.0209} & 0.9983\,{\scriptsize$\pm$\,.0013} & 0.9983\,{\scriptsize$\pm$\,.0096} & 0.9983\,{\scriptsize$\pm$\,.0370} \\
     &  & Thompson & 0.9976\,{\scriptsize$\pm$\,.0002} & 0.9985\,{\scriptsize$\pm$\,.0561} & 0.9969\,{\scriptsize$\pm$\,.0100} & 0.9967\,{\scriptsize$\pm$\,.0827} & 0.6872\,{\scriptsize$\pm$\,.0873} \\
    \cmidrule(lr){2-8}
     & \multirow{3}{*}{MMD} & UCB & 0.7172\,{\scriptsize$\pm$\,.0005} & \textbf{0.9603\,{\scriptsize$\pm$\,.0312}} & \textbf{0.9603\,{\scriptsize$\pm$\,.0129}} & \textbf{0.9603\,{\scriptsize$\pm$\,.0124}} & 0.8634\,{\scriptsize$\pm$\,.0000} \\
     &  & EXP3.P & 0.7172\,{\scriptsize$\pm$\,.0005} & \textbf{0.9603\,{\scriptsize$\pm$\,.0394}} & 0.9595\,{\scriptsize$\pm$\,.0386} & 0.9603\,{\scriptsize$\pm$\,.0159} & 0.9603\,{\scriptsize$\pm$\,.0124} \\
     &  & Thompson & 0.7172\,{\scriptsize$\pm$\,.0005} & 0.9521\,{\scriptsize$\pm$\,.1024} & 0.9603\,{\scriptsize$\pm$\,.0546} & 0.9603\,{\scriptsize$\pm$\,.0016} & 0.9603\,{\scriptsize$\pm$\,.0000} \\
    \cmidrule(lr){2-8}
     & \multirow{3}{*}{HT} & UCB & \textbf{0.8104\,{\scriptsize$\pm$\,.0355}} & \textbf{0.8104\,{\scriptsize$\pm$\,.0651}} & 0.7250\,{\scriptsize$\pm$\,.0684} & 0.7250\,{\scriptsize$\pm$\,.0994} & 0.6347\,{\scriptsize$\pm$\,.1377} \\
     &  & EXP3.P & 0.7956\,{\scriptsize$\pm$\,.0274} & \textbf{0.8104\,{\scriptsize$\pm$\,.0942}} & 0.7421\,{\scriptsize$\pm$\,.1000} & 0.7250\,{\scriptsize$\pm$\,.0971} & 0.7250\,{\scriptsize$\pm$\,.1502} \\
     &  & Thompson & \textbf{0.8104\,{\scriptsize$\pm$\,.0158}} & \textbf{0.8104\,{\scriptsize$\pm$\,.1471}} & \textbf{0.8104\,{\scriptsize$\pm$\,.1785}} & 0.7933\,{\scriptsize$\pm$\,.1707} & 0.7250\,{\scriptsize$\pm$\,.1571} \\
    \midrule
    \multirow{6}{*}{Qwen3-14B} & \multirow{3}{*}{CP} & UCB & 0.9579\,{\scriptsize$\pm$\,.0012} & 0.9972\,{\scriptsize$\pm$\,.0038} & 0.9972\,{\scriptsize$\pm$\,.0074} & 0.9938\,{\scriptsize$\pm$\,.0197} & 0.9929\,{\scriptsize$\pm$\,.0120} \\
     &  & EXP3.P & 0.9847\,{\scriptsize$\pm$\,.0222} & 0.9971\,{\scriptsize$\pm$\,.0054} & 0.9967\,{\scriptsize$\pm$\,.0114} & 0.9926\,{\scriptsize$\pm$\,.0178} & 0.9817\,{\scriptsize$\pm$\,.0175} \\
     &  & Thompson & \textbf{1.0003\,{\scriptsize$\pm$\,.0199}} & 0.9887\,{\scriptsize$\pm$\,.0070} & \textbf{1.0003\,{\scriptsize$\pm$\,.0143}} & \textbf{1.0003\,{\scriptsize$\pm$\,.0042}} & \textbf{1.0003\,{\scriptsize$\pm$\,.0058}} \\
    \cmidrule(lr){2-8}
     & \multirow{3}{*}{MMD} & UCB & 0.9928\,{\scriptsize$\pm$\,.0537} & 0.9603\,{\scriptsize$\pm$\,.0327} & 0.9955\,{\scriptsize$\pm$\,.0171} & 0.8828\,{\scriptsize$\pm$\,.0098} & 0.8634\,{\scriptsize$\pm$\,.0223} \\
     &  & EXP3.P & 0.9745\,{\scriptsize$\pm$\,.0482} & 0.9630\,{\scriptsize$\pm$\,.0454} & 0.9809\,{\scriptsize$\pm$\,.0346} & 0.9953\,{\scriptsize$\pm$\,.0471} & 0.9447\,{\scriptsize$\pm$\,.0530} \\
     &  & Thompson & 0.8989\,{\scriptsize$\pm$\,.0395} & \textbf{0.9970\,{\scriptsize$\pm$\,.0410}} & 0.9968\,{\scriptsize$\pm$\,.0164} & 0.9953\,{\scriptsize$\pm$\,.0240} & 0.9947\,{\scriptsize$\pm$\,.0142} \\
    \midrule
    \multirow{9}{*}{Llama} & \multirow{3}{*}{CP} & UCB & 0.9259\,{\scriptsize$\pm$\,.0083} & 0.9458\,{\scriptsize$\pm$\,.0679} & 0.9451\,{\scriptsize$\pm$\,.0633} & 0.9259\,{\scriptsize$\pm$\,.2337} & 0.9259\,{\scriptsize$\pm$\,.0253} \\
     &  & EXP3.P & 0.9285\,{\scriptsize$\pm$\,.0080} & 0.9480\,{\scriptsize$\pm$\,.0687} & 0.9484\,{\scriptsize$\pm$\,.0716} & 0.9259\,{\scriptsize$\pm$\,.1632} & 0.9468\,{\scriptsize$\pm$\,.0238} \\
     &  & Thompson & 0.9259\,{\scriptsize$\pm$\,.0083} & \textbf{0.9533\,{\scriptsize$\pm$\,.0689}} & \textbf{0.9533\,{\scriptsize$\pm$\,.0706}} & 0.9259\,{\scriptsize$\pm$\,.1767} & 0.9433\,{\scriptsize$\pm$\,.0270} \\
    \cmidrule(lr){2-8}
     & \multirow{3}{*}{MMD} & UCB & 0.8487\,{\scriptsize$\pm$\,.2946} & 0.7978\,{\scriptsize$\pm$\,.0362} & 0.7978\,{\scriptsize$\pm$\,.0575} & \textbf{0.8614\,{\scriptsize$\pm$\,.0686}} & 0.7978\,{\scriptsize$\pm$\,.0689} \\
     &  & EXP3.P & 0.8487\,{\scriptsize$\pm$\,.2946} & 0.7978\,{\scriptsize$\pm$\,.0057} & 0.7978\,{\scriptsize$\pm$\,.0542} & \textbf{0.8614\,{\scriptsize$\pm$\,.1270}} & 0.8232\,{\scriptsize$\pm$\,.0283} \\
     &  & Thompson & 0.8487\,{\scriptsize$\pm$\,.2946} & 0.7978\,{\scriptsize$\pm$\,.0089} & 0.7978\,{\scriptsize$\pm$\,.0448} & \textbf{0.8614\,{\scriptsize$\pm$\,.1731}} & 0.7978\,{\scriptsize$\pm$\,.0143} \\
    \cmidrule(lr){2-8}
     & \multirow{3}{*}{HT} & UCB & 0.0000\,{\scriptsize$\pm$\,.0000} & 0.0000\,{\scriptsize$\pm$\,.0000} & 0.0000\,{\scriptsize$\pm$\,.0000} & 0.0000\,{\scriptsize$\pm$\,.0000} & 0.0000\,{\scriptsize$\pm$\,.0000} \\
     &  & EXP3.P & 0.0000\,{\scriptsize$\pm$\,.0001} & 0.0000\,{\scriptsize$\pm$\,.0000} & 0.0000\,{\scriptsize$\pm$\,.0000} & 0.0000\,{\scriptsize$\pm$\,.0000} & 0.0000\,{\scriptsize$\pm$\,.0000} \\
     &  & Thompson & 0.0000\,{\scriptsize$\pm$\,.0001} & 0.0000\,{\scriptsize$\pm$\,.0000} & 0.0000\,{\scriptsize$\pm$\,.0000} & 0.0000\,{\scriptsize$\pm$\,.0000} & 0.0000\,{\scriptsize$\pm$\,.0000} \\
    \bottomrule
    \end{tabular}
    \end{adjustbox}
\end{table*}

\tcbset{
  promptbox/.style={
    breakable, enhanced,
    colback=gray!4, colframe=gray!55!black,
    boxrule=0.5pt, left=6pt, right=6pt, top=4pt, bottom=4pt,
    fonttitle=\bfseries\small, coltitle=white,
    colbacktitle=gray!55!black,
    boxed title style={boxrule=0pt},
    before skip=6pt, after skip=8pt,
    listing only,
    listing options={
      basicstyle=\ttfamily\footnotesize,
      breaklines=true,
      breakatwhitespace=false,
      breakindent=0pt,
      columns=flexible,
      keepspaces=true,
      upquote=true,
      showstringspaces=false,
      aboveskip=0pt, belowskip=0pt,
    },
  }
}

\section{Extension Experiments: LLM Prompt Optimization (LPO)}
\label{app:lpo}
Beyond three geometric optimization coding problems, we deploy one more task, namely, LLM Prompt Optimization (LPO), from OpenEvolve, where the evolving target is a natural-language prompt, and the fitness is evaluated by the LLM's answer accuracy on HotpotQA questions using Qwen3-1.7B. As shown in \Cref{tab:lpo}, all regularities introduced in the main paper apply: the envelope applies ($R^2=0.92$); the bilinear form fits; and the depth gain is small but statistically detectable (+0.028, p=0.047), consistent with the capability gate for a 1.7B mutator near a high best-of-N baseline, demonstrating the potential of generalizing our bilinear regularities to a cross-domain end-to-end scenarios.

\begin{table}[h!]
\caption{LLM Prompt optimization (LPO) on the first 40 HotpotQA questions with Qwen3-1.7B at $C=128$: both regularities and the capability gate replicate. Envelope $R^2$ is the fit of $V_{\max}$ against effective FLOPs (\Cref{sec:envelope}); the bilinear
coefficients are for \Cref{eq:bi-law}; $V_{T=1}$ is the best-of-$N$ fitness, $V_{\max}$ the best over all depths,
$\mathrm{pen}_{\mathrm{BoN}}=V_{\max}-V_{T=1}$ the depth gain, and $p$ its permutation test (\Cref{app:effect_model_size}). The fitness score of LPO is the answer accuracy rate on the HotpotQA pool.}
\label{tab:lpo}
\centering
\small
\begin{tabular}{llr}
\toprule
Analysis & Statistic & Value \\
\midrule
\multirow{1}{*}{Envelope (\Cref{sec:envelope})}
  & $R^2$ & 0.923 \\
\midrule
\multirow{5}{*}{Bilinear fit}
  & $\beta_0$ & $-1.599$ \\
  & $a$       & $-0.108$ \\
  & $b$       & $-0.105$ \\
  & $c$       & $-0.005$ \\
  & $R^2$     & \phantom{$-$}0.732 \\
\midrule
\multirow{4}{*}{Capability gate}
  & $V_{T=1}$                        & 0.872 \\
  & $V_{\max}$                       & 0.900 \\
  & $\text{pen}_{\text{BoN}}$        & $+0.028$ \\
  & $p$                              & 0.047 \\
\bottomrule
\end{tabular}
\end{table}

\section{Prompts}
\label{app:prompts}

This appendix lists, verbatim, the system prompts used by the three
\emph{*Evolve} frameworks compared in the paper --- OpenEvolve,
CodeEvolve, and ShinkaEvolve --- on the three tasks we study: Circle
Packing (CP, $n=26$ in the unit square), Min/Max Distance (MMD, $n=16$
in 2D), and Heilbronn Triangle (HT, $n=11$ in an equilateral triangle).
Each task is specified by a fixed objective, a benchmark constant from
AlphaEvolve, and a constructor function with a fixed name and return
signature; these are shared across all three frameworks. What differs
between frameworks is the \emph{scaffold} around the task: the chat
format, what context from the archive is surfaced to the LLM, and the
output protocol the LLM is asked to follow (full rewrite vs.\
SEARCH/REPLACE edit, free-form vs.\ tagged response). The subsections
below give each framework's prompts in turn.

\subsection{OpenEvolve (also used by Greedy and \texttt{BaSE})}
\label{app:prompts:openevolve}

Our Greedy baseline and the proposed \texttt{BaSE} allocator use the OpenEvolve
prompts \emph{verbatim} (the same system message, user-message
template, and per-task instructions reproduced below) so that any
fitness difference between Greedy, \texttt{BaSE}, and OpenEvolve reflects
\emph{allocation} of the LLM call budget rather than prompt engineering.
OpenEvolve composes each call as one system message and one user
message. We use the full-rewrite mode with three top programs and two
diverse programs from the archive surfaced in each call. The user
message wraps the per-task system message below together with the
current program, its fitness, MAP-Elites feature coordinates, and the
top and diverse program archive.

\begin{tcblisting}{promptbox, title=OpenEvolve: user-message template}
# Current Program Information
- Fitness: {fitness_score}
- Feature coordinates: {feature_coords}
- Focus areas: {improvement_areas}

{artifacts}

# Program Evolution History
{evolution_history}

# Current Program
```{language}
{current_program}
```

# Task
Rewrite the program to improve its FITNESS SCORE.
The system maintains diversity across these dimensions: {feature_dimensions}
Different solutions with similar fitness but different features are valuable.
Provide the complete new program code.

IMPORTANT: Make sure your rewritten program maintains the same inputs and
outputs as the original program, but with improved internal implementation.

```{language}
# Your rewritten program here
```
\end{tcblisting}

\begin{tcblisting}{promptbox, title=OpenEvolve: CP system message}
You are an expert mathematician specializing in circle packing problems and
computational geometry. Your task is to improve a constructor function that
directly produces a specific arrangement of 26 circles in a unit square,
maximizing the sum of their radii. The AlphaEvolve paper achieved a sum of
2.635 for n=26.

Key geometric insights:
- Circle packings often follow hexagonal patterns in the densest regions
- Maximum density for infinite circle packing is pi/(2*sqrt(3)) ~= 0.9069
- Edge effects make square container packing harder than infinite packing
- Circles can be placed in layers or shells when confined to a square
- Similar radius circles often form regular patterns, while varied radii allow
  better space utilization
- Perfect symmetry may not yield the optimal packing due to edge effects

Focus on designing an explicit constructor that places each circle in a
specific position, rather than an iterative search algorithm.
\end{tcblisting}

\begin{tcblisting}{promptbox, title=OpenEvolve: MMD system message}
You are an expert computational geometer and optimization specialist focusing
on point dispersion problems.
Your task is to improve a constructor function that generates an optimal
arrangement of exactly 16 points in 2D space, maximizing the ratio of minimum
distance to maximum distance between all point pairs.

PROBLEM CONTEXT:
- Target: Beat the AlphaEvolve benchmark of min/max ratio
  = 1/sqrt(12.889266112) ~= 0.2786
- Mathematical formulation: For points Pi = (xi, yi), i = 1,...,16:
  * Distance matrix: dij = sqrt[(xi-xj)^2 + (yi-yj)^2] for all i!=j
  * Minimum distance: dmin = min{dij : i!=j}
  * Maximum distance: dmax = max{dij : i!=j}
  * Objective: maximize (dmin/dmax)^2
- The metric is inv_ratio_squared = (dmin/dmax)^2 and
  combined_score = inv_ratio_squared / BENCHMARK
- BENCHMARK = 1/12.889266112

STRATEGIES TO CONSIDER:
- Regular polygon arrangements (vertices of regular n-gon)
- Multiple concentric rings of points
- Hexagonal lattice subsets
- Numerical optimization (scipy.optimize) to fine-tune positions
- Gradient descent on the min-distance objective

The function must be named min_max_dist_dim2_16() and return a numpy array
of shape (16, 2).
\end{tcblisting}

\begin{tcblisting}{promptbox, title=OpenEvolve: HT system message}
You are an expert computational geometer and optimization specialist focusing
on point-dispersion problems in convex regions.
Your task is to improve a constructor function that places exactly 11 points
inside an equilateral triangle to maximize the area of the smallest triangle
formed by any three of the 11 points (the Heilbronn problem for triangles).

PROBLEM CONTEXT:
- Container: equilateral triangle with vertices (0,0), (1,0), (0.5, sqrt(3)/2).
  Area = sqrt(3)/4.
- Objective: maximize min_area_normalized = min{Area(P_i, P_j, P_k)} /
  Area(container), over all C(11,3)=165 triples.
- Target: beat the AlphaEvolve benchmark of
  min_area_normalized = 0.036529889880030156.
- The metric is combined_score = min_area_normalized / BENCHMARK. Goal: > 1.0.

STRATEGIES TO CONSIDER:
- Symmetric arrangements that exploit the triangle's three-fold symmetry
  (vertex configurations + interior layouts).
- Place a subset of points on or near the boundary (vertices + edge midpoints
  + sub-divisions), with the remainder in the interior on rings.
- Numerical refinement: scipy.optimize.minimize on the 22 coordinate variables,
  with the objective `max(-min_triangle_area)` or smoothed log-sum-exp
  surrogate.
- Lattice / hexagonal sub-patterns inside the triangle.
- Random multistart followed by local refinement to escape symmetry-broken
  local optima.

TECHNICAL REQUIREMENTS:
- Determinism: if you use randomness, fix the seed (e.g., numpy.random.seed(42)).
- Output: return an (11, 2) numpy.ndarray of float (x, y) coordinates, all
  strictly inside the triangle (tolerance 1e-6).
- The function MUST be named heilbronn_triangle11() and take no arguments.
\end{tcblisting}

\subsection{CodeEvolve}
\label{app:prompts:codeevolve}

CodeEvolve uses a multi-turn chat format that replays the program's parent
lineage as alternating user and assistant turns. The system message is
assembled as: the per-task system block below, followed by a
computational-budget block, followed by one of four task templates
(exploitation or exploration, each with or without inspirations). All
four templates require the LLM to emit changes in strict
\texttt{<<<<<<< SEARCH ... ======= ... >>>>>>> REPLACE} form inside
designated edit regions; full rewrites are not supported.

\begin{tcblisting}{promptbox, title=CodeEvolve: exploitation task template (no inspirations)}
# TASK: CODE EVOLUTION
Your goal is to evolve the provided program by modifying specific sections.
You MUST adhere strictly to the SEARCH/REPLACE format described below for
all modifications.

## MODIFICATION FORMAT:
Present your proposed code changes using the following structure:
    ```
    <<<<<<< SEARCH
    [exact original code STRICTLY WITHIN an EVOLVE-BLOCK]
    =======
    [your modified code]
    >>>>>>> REPLACE
    ```
* For multiple independent changes, provide each in a separate SEARCH/REPLACE
  block.

## CORE RULES FOR CODE MODIFICATION:

### Scope & Boundaries:
    1. Target EVOLVE-BLOCK ONLY: All code modifications MUST be confined to
       sections explicitly marked between EVOLVE-BLOCK-START and
       EVOLVE-BLOCK-END comments.
    2. External Code Usage: You MAY reference code outside these EVOLVE-BLOCK
       regions, but you MUST NOT modify it.
    3. New Imports: If new imports are required, add them within an
       EVOLVE-BLOCK.

### SEARCH Block Requirements:
    1. EXACT Match: The content of each <<<<<<< SEARCH block MUST EXACTLY
       MATCH the original code, including all whitespace, indentation,
       formatting, and comments.
    2. No Comment Alterations in SEARCH: Do NOT add, remove, or modify
       comments within the SEARCH block.
    3. First Occurrence Precedence: If multiple identical code sections exist
       in the original program, your SEARCH block will be applied to the
       first occurrence matching its content.

### Output & Compatibility:
    1. Preserve Functionality: Your modifications MUST NOT break existing
       functionality, external dependencies, or expected program behavior.
    2. Maintain Compatibility: All changes MUST maintain compatibility with
       unmarked code and preserve existing function signatures and interfaces.
    3. Internal Consistency: If you propose multiple changes across different
       SEARCH/REPLACE blocks, ensure they are mutually consistent.

## COMPUTATIONAL BUDGET:
- Time limit: {timeout_s} seconds maximum execution time
- Memory limit: {max_mem}
\end{tcblisting}

The exploration variant retitles the task as code exploration and
diversification, asking for novel strategies and distinct algorithmic
pathways rather than incremental refinement. The with-inspirations
variants additionally require the LLM to analyse the supplied inspiration
programs and synthesise their differences before producing the
SEARCH/REPLACE block.

\begin{tcblisting}{promptbox, title=CodeEvolve: CP system block}
You are an expert computational geometer and optimization specialist focusing
on circle-packing problems.
Your task is to improve a constructor function that arranges exactly 26
non-overlapping circles inside the unit square [0,1]x[0,1], maximizing the
sum of their radii.

PROBLEM CONTEXT:
- Target: approach or beat the AlphaEvolve benchmark of sum_radii = 2.635
  for n=26
- Container: unit square with side length 1
- Constraints:
  * All 26 circles must be fully contained in [0,1]x[0,1]
  * No two circles may overlap
  * All radii must be positive and finite
- The metric is combined_score = sum_radii / 2.635

STRATEGIES TO CONSIDER:
- Hexagonal or honeycomb placement with edge-row adjustments
- Concentric shells or rings with varying radii
- Asymmetric layouts that exploit square-boundary effects
- Post-placement scaling so adjacent circles touch exactly
- Numerical refinement of center positions after seeding a pattern

The evolve target is construct_packing(), which must return (centers, radii,
sum_radii), where centers has shape (26, 2), radii has shape (26,), and
sum_radii is a scalar.
\end{tcblisting}

\begin{tcblisting}{promptbox, title=CodeEvolve: MMD system block}
You are an expert computational geometer and optimization specialist focusing
on point dispersion problems.
Your task is to improve a constructor function that generates an optimal
arrangement of exactly 16 points in 2D space, maximizing the ratio of minimum
distance to maximum distance between all point pairs.

PROBLEM CONTEXT:
- Target: Beat the AlphaEvolve benchmark of min/max ratio
  = 1/sqrt(12.889266112) ~= 0.2786
- Mathematical formulation: For points Pi = (xi, yi), i = 1,...,16:
  * Distance matrix: dij = sqrt[(xi-xj)^2 + (yi-yj)^2] for all i!=j
  * Minimum distance: dmin = min{dij : i!=j}
  * Maximum distance: dmax = max{dij : i!=j}
  * Objective: maximize (dmin/dmax)^2
- The metric is combined_score = ((dmin/dmax)^2) / (1/12.889266112)

STRATEGIES TO CONSIDER:
- Regular polygon arrangements
- Multiple concentric rings of points
- Hexagonal lattice subsets
- Numerical optimization with scipy.optimize
- Gradient-based or derivative-free refinement of point coordinates

The function must be named min_max_dist_dim2_16() and return a numpy array
of shape (16, 2).
\end{tcblisting}

\begin{tcblisting}{promptbox, title=CodeEvolve: HT system block}
You are an expert computational geometer and optimization specialist focusing
on point-dispersion problems in convex regions.
Your task is to improve a constructor function that places exactly 11 points
inside an equilateral triangle to maximize the area of the smallest triangle
formed by any three of the 11 points (the Heilbronn problem for triangles).

PROBLEM CONTEXT:
- Container: equilateral triangle with vertices (0,0), (1,0), (0.5, sqrt(3)/2).
  Area = sqrt(3)/4.
- Objective: maximize min_area_normalized = min{Area(P_i, P_j, P_k)} /
  Area(container), over all C(11,3)=165 triples.
- Target: beat the AlphaEvolve benchmark of
  min_area_normalized = 0.036529889880030156.
- The metric is combined_score = min_area_normalized / BENCHMARK. Goal: > 1.0.

STRATEGIES TO CONSIDER:
- Symmetric arrangements that exploit the triangle's three-fold symmetry
  (vertex configurations + interior layouts).
- Place a subset of points on or near the boundary (vertices + edge midpoints
  + sub-divisions), with the remainder in the interior on rings.
- Numerical refinement: scipy.optimize.minimize on the 22 coordinate variables,
  with the objective `max(-min_triangle_area)` or smoothed log-sum-exp
  surrogate.
- Lattice / hexagonal sub-patterns inside the triangle.
- Random multistart followed by local refinement to escape symmetry-broken
  local optima.

TECHNICAL REQUIREMENTS:
- Determinism: if you use randomness, fix the seed (e.g., numpy.random.seed(42)).
- Output: return an (11, 2) numpy.ndarray of float (x, y) coordinates, all
  strictly inside the triangle (tolerance 1e-6).
- The function MUST be named heilbronn_triangle11() and take no arguments.
\end{tcblisting}

\subsection{ShinkaEvolve}
\label{app:prompts:shinka}

ShinkaEvolve concatenates the per-task system prompt below with one of
five full-rewrite format variants sampled uniformly per call (default,
different-algorithm, context-motivated, structural-redesign, and
parametric-design), and pairs it with an iteration message that carries
the current program and its metrics.

\begin{tcblisting}{promptbox, title=ShinkaEvolve: full-rewrite format suffix}
Rewrite the program to improve its performance on the specified metrics.
Provide the complete new program code.
You MUST respond using a short summary name, description and the full code:

<NAME>
A shortened name summarizing the code you are proposing. Lowercase, no spaces,
underscores allowed.
</NAME>

<DESCRIPTION>
A description and argumentation process of the code you are proposing.
</DESCRIPTION>

<CODE>
```{language}
# The new rewritten program here.
```
</CODE>

* Keep the markers "EVOLVE-BLOCK-START" and "EVOLVE-BLOCK-END" in the code.
  Do not change the code outside of these markers.
* Make sure your rewritten program maintains the same inputs and outputs as
  the original program, but with improved internal implementation.
* Make sure the file still runs after your changes.
* Use the <NAME>, <DESCRIPTION>, and <CODE> delimiters to structure your
  response. It will be parsed afterwards.
\end{tcblisting}

\begin{tcblisting}{promptbox, title=ShinkaEvolve: iteration message}
# Current program

Here is the current program we are trying to improve (you will need to
propose a new program with the same inputs and outputs as the original
program, but with improved internal implementation):

```{language}
{code_content}
```

Here are the performance metrics of the program:

{performance_metrics}{text_feedback_section}

# Task
Rewrite the program to improve its performance on the specified metrics.
Provide the complete new program code.

IMPORTANT: Make sure your rewritten program maintains the same inputs and
outputs as the original program, but with improved internal implementation.
\end{tcblisting}

\begin{tcblisting}{promptbox, title=ShinkaEvolve: CP system prompt}
You are an expert computational geometer and optimization specialist focusing
on circle-packing problems.
Your task is to improve a constructor function that arranges exactly 26
non-overlapping circles inside the unit square [0,1]x[0,1], maximizing the
sum of their radii.

PROBLEM CONTEXT:
- Target: approach or beat the AlphaEvolve benchmark of sum_radii = 2.635
  for n=26
- Container: unit square with side length 1
- Constraints:
  * All 26 circles must be fully contained in [0,1]x[0,1]
  * No two circles may overlap
  * All radii must be positive and finite
- The metric is combined_score = sum_radii / 2.635

STRATEGIES TO CONSIDER:
- Hexagonal or honeycomb placement with edge-row adjustments
- Concentric shells or rings with varying radii
- Asymmetric layouts that exploit square-boundary effects
- Post-placement scaling so adjacent circles touch exactly
- Numerical refinement (scipy.optimize) on center positions after seeding
  a pattern

The evolve target is construct_packing(), which must return (centers, radii,
sum_radii), where centers has shape (26, 2), radii has shape (26,), and
sum_radii is a scalar.
\end{tcblisting}

\begin{tcblisting}{promptbox, title=ShinkaEvolve: MMD system prompt}
You are an expert computational geometer and optimization specialist focusing
on point dispersion problems.
Your task is to improve a constructor function that generates an optimal
arrangement of exactly 16 points in 2D space, maximizing the ratio of minimum
distance to maximum distance between all point pairs.

PROBLEM CONTEXT:
- Target: Beat the AlphaEvolve benchmark of min/max ratio
  = 1/sqrt(12.889266112) ~= 0.2786
- Mathematical formulation: For points Pi = (xi, yi), i = 1,...,16:
  * Distance matrix: dij = sqrt[(xi-xj)^2 + (yi-yj)^2] for all i!=j
  * Minimum distance: dmin = min{dij : i!=j}
  * Maximum distance: dmax = max{dij : i!=j}
  * Objective: maximize (dmin/dmax)^2
- The metric is inv_ratio_squared = (dmin/dmax)^2 and
  combined_score = inv_ratio_squared / BENCHMARK
- BENCHMARK = 1/12.889266112

STRATEGIES TO CONSIDER:
- Regular polygon arrangements (vertices of regular n-gon)
- Multiple concentric rings of points
- Hexagonal lattice subsets
- Numerical optimization (scipy.optimize) to fine-tune positions
- Gradient descent on the min-distance objective

The function must be named min_max_dist_dim2_16() and return a numpy array
of shape (16, 2).
\end{tcblisting}

\begin{tcblisting}{promptbox, title=ShinkaEvolve: HT system prompt}
You are an expert computational geometer and optimization specialist focusing
on the Heilbronn triangle problem.
Your task is to improve a constructor function that places exactly 11 points
inside (or on the boundary of) an equilateral triangle with vertices (0,0),
(1,0), and (0.5, sqrt(3)/2), so as to maximize the area of the smallest
triangle formed by any three of the points.

PROBLEM CONTEXT:
- Target: approach or beat the AlphaEvolve benchmark of
  min_area_normalized = 0.036529889880030156 for n=11
- Container: equilateral triangle with vertices (0,0), (1,0), (0.5, sqrt(3)/2).
  Its area is sqrt(3)/4.
- Constraints:
  * All 11 points must lie inside or on the boundary of the triangle (within
    tolerance 1e-6)
  * The objective is to maximize min{ area(p_i, p_j, p_k) : 1 <= i < j < k <= 11 }
  * The reported "min_area_normalized" is that minimum triangle area divided
    by the host triangle's area (sqrt(3)/4)
- The metric is combined_score = min_area_normalized / 0.036529889880030156

CRITICAL - degeneracy warning:
- combined_score = 0 if ANY three of the 11 points are collinear (the smallest
  triangle's area is then 0).
- Regular lattices, points along a triangle edge, and rotationally-symmetric
  grids almost always contain 3 collinear points and therefore score 0. Avoid
  them, or perturb them after seeding so no three points share a line.
- If you use a structured seed, follow it with numerical refinement (e.g.
  scipy.optimize) on the actual min-triangle-area (or a soft-min surrogate)
  to break collinearity.

The function must be named heilbronn_triangle11() and return a numpy array
of shape (11, 2).
\end{tcblisting}

\section{Use of AI Assistants}
\label{app:ai}

We used large language models (Anthropic's Claude) to assist with 
paper revising.

\section{Artifact Licenses}
\label{app:licenses}

We use the following artifacts under their respective licenses, 
consistent with their intended research use:
\begin{itemize}
  \item \textbf{OpenEvolve} (Apache-2.0): we use the example 
        evaluators, initial programs, and prompt templates for 
        Circle Packing, MinMaxDist, and Heilbronn Triangle verbatim.
  \item \textbf{Qwen3 1.7B / 4B / 8B / 14B} (Apache-2.0): used as 
        the mutation engine in inference mode only; no weights 
        modified.
  \item \textbf{Llama-3.1-8B} (Llama 3.1 Community License): used 
        as the mutation engine in inference mode only; no weights 
        modified.
  \item \textbf{vLLM} (Apache-2.0): used as the inference server.
\end{itemize}

\section{Ethics Statement}

This work uses publicly available open-weight LLMs (Qwen3, 
Llama-3.1) and public geometric-optimization benchmarks. It 
involves no human subjects, no personal data, and no deployment 
in safety-critical settings. We foresee no direct ethical risks.

\section{Potential Risks}

LLM-generated programs may contain correctness or security flaws 
that fitness-based evaluators do not catch, and adaptive 
allocation can concentrate compute on superficially promising 
trajectories.

\end{document}